\documentclass[letterpaper, 10 pt, conference]{ieeeconf}  % Comment this line out if you need a4paper

\IEEEoverridecommandlockouts                              % This command is only needed if 
\usepackage[utf8]{inputenc} % allow utf-8 input
\usepackage[T1]{fontenc}    % use 8-bit T1 fonts
\usepackage{url}            % simple URL typesetting
\usepackage{booktabs}       % professional-quality tables
\usepackage{amsfonts}       % blackboard math symbols
\usepackage{microtype}      % microtypography

\usepackage{amsmath}
\usepackage{amssymb}
\usepackage{mathtools}
\usepackage{tensor}
\usepackage{bm}  % assumes amsmath package installed
\usepackage{graphics} % for pdf, bitmapped graphics files
\usepackage{epsfig} % for postscript graphics files
\usepackage{wrapfig}
\usepackage{subcaption}
\usepackage[font=small,labelfont=small]{caption}
\usepackage[pagebackref=true]{hyperref}
\usepackage{color}
\usepackage{algorithm}
\usepackage{algorithmic}
\usepackage{sidecap}
\sidecaptionvpos{figure}{c}
\usepackage{tikz}
\usetikzlibrary{arrows.meta}
\usepackage{tabularx}
\usepackage{colortbl}
\usepackage{booktabs}
\usepackage{multirow}
\usepackage{siunitx}
\usepackage{nicefrac}
\usepackage[nolist]{acronym} % acronyms by the \ac{label} command
\usepackage{cuted}

\graphicspath{{figures/}}

\definecolor{lightred}{rgb}{0.8, 0.22, 0.29}
\definecolor{turquoise}{rgb}{0.25, 0.89, 0.82}
\definecolor{mediumorchid}{rgb}{0.73, 0.33, 0.83}
\definecolor{darkorange}{rgb}{1.0, 0.65, 0.0}
\definecolor{marroon}{rgb}{0.75, 0.3, 0.0}
\definecolor{navy}{rgb}{0.0, 0.0, 0.5}
\definecolor{dodgerblue}{rgb}{0.12, 0.565, 1.0}
\definecolor{darkdodgerblue}{rgb}{0.06, 0.4, 0.8}
\definecolor{crimson}{rgb}{0.86, 0.08, 0.235}
\definecolor{lightgray}{rgb}{0.6, 0.6, 0.6}
\definecolor{darkgreen}{rgb}{0.0, 0.39, 0.0}
\definecolor{middlegreen}{rgb}{0.0, 0.5, 0.0}
\definecolor{middleyellow}{rgb}{1.0, 0.7, 0.0}
\definecolor{aquamarine}{rgb}{0.5, 1.0, 0.83}
\definecolor{lightblue}{rgb}{0.5, 0.7, 1.0}
\definecolor{olive}{rgb}{0.5, 0.5, 0.0}
\definecolor{lightgray}{rgb}{0.6, 0.6, 0.6}

\DeclareRobustCommand{\crimsoncircle}{\tikz{ \filldraw[color=white, fill=crimson, thick](0,0) circle (.075);}}

\DeclareRobustCommand{\middlegreencircle}{\tikz{ \filldraw[color=white, fill=middlegreen, thick](0,0) circle (.075);}}

\DeclareRobustCommand{\marrooncircle}{\tikz{ \filldraw[color=white, fill=marroon, thick](0,0) circle (.075);}}

\DeclareRobustCommand{\olivecircle}{\tikz{ \filldraw[color=white, fill=olive, thick](0,0) circle (.075);}}
\DeclareRobustCommand{\lightgraycircle}{\tikz{ \filldraw[color=white, fill=lightgray, thick](0,0) circle (.075);}}

\DeclareRobustCommand{\marroonline}{\raisebox{2pt}{\tikz{\draw[marroon,solid,line width = 1.1pt](0,0) -- (4mm,0);}}}
\DeclareRobustCommand{\crimsonline}{\raisebox{2pt}{\tikz{\draw[crimson,solid,line width = 1.1pt](0,0) -- (4mm,0);}}}

\DeclareRobustCommand{\blackline}{\raisebox{2pt}{\tikz{\draw[black,solid,line width = 1.1pt](0,0) -- (4mm,0);}}}

\DeclareRobustCommand{\blackdashedline}{\raisebox{2pt}{\tikz{\draw[black,dashed,line width = 1.1pt](0,0) -- (4mm,0);}}}
\DeclareRobustCommand{\middlegreenline}{\raisebox{2pt}{\tikz{\draw[middlegreen,solid,line width = 1.1pt](0,0) -- (4mm,0);}}}

\DeclareRobustCommand{\oliveline}{\raisebox{2pt}{\tikz{\draw[olive,solid,line width = 1.1pt](0,0) -- (4mm,0);}}}

\newcommand{\etal}{\MakeLowercase{\textit{et al.}}}
\newcommand{\trsp}{\mathsf{T}}

\DeclareMathOperator*{\argmax}{argmax} % thin space, limits underneath in displays
\DeclareSymbolFont{bbold}{U}{bbold}{m}{n}
\DeclareSymbolFontAlphabet{\mathbbold}{bbold}

\newcommand{\euclideanspace}{\mathbb{R}}
\newcommand{\manifold}{\mathcal{M}}

\newcommand{\innerprod}[3]{\langle #2, #3 \rangle_{#1}}  % Inner product of #2 and #3 at #1
\newcommand{\norm}[2]{\| #2\|_{#1}}  % Norm of #2 at #1
\newcommand{\expmapblank}[1]{\text{Exp}_{#1}}  % Exponential map of #2 at #1
\newcommand{\expmap}[2]{\expmapblank{#1}(#2)}  % Exponential map of #2 at #1
\newcommand{\logmap}[2]{\text{Log}_{#1}(#2)}  % Logarithmic map of #2 at #1
\newcommand{\lorentz}[1]{\mathbb{H}^{#1}_{\mathcal{L}}} % Notation of hyperbolic manifold of dimensionality #1
\newcommand{\tangentspacelorentz}[2]{\mathcal{T}_{#1}\lorentz{#2}}  % Tangent space for hyperbolic manifold
\newcommand{\poincare}[1]{\mathbb{H}^{#1}_{\mathcal{P}}} % Notation of Poincaré model of hyperbolic manifold of dimensionality #1
\newcommand{\manifolddist}[2]{d_{\manifold}(#1, #2)}
\newcommand{\hypedist}[2]{d_{\lorentz{D_x}}(#1, #2)}
\newcommand{\graphdist}[2]{d_{\mathbb{G}}(#1, #2)}
\newcommand{\hypenormallatent}[3]{\mathcal{N}_{\lorentz{D_x}}(#1;#2,#3)} % Hyperbolic Gaussian with sample #1, mean #2 and covariance #3
\newcommand{\gaussiandist}[3]{\mathcal{N}(#1;#2,#3)} % Gaussian with sample #1, mean #2 and covariance #3
\newcommand{\proj}{\mathrm{proj}}

\newcommand{\metric}{\bm{G}}
\newcommand{\lorentzmetric}{\bm{G}^{\mathcal{L}}}
\newcommand{\pullbackmetric}{\bm{G}^{\text{P}}}
\newcommand{\euclpullbackmetric}{\bm{G}^{\text{P},\euclideanspace}}
\newcommand{\lorentzpullbackmetric}{\bm{G}^{\text{P},\mathcal{L}}}

\newcommand{\localcord}[1]{\textcolor{darkdodgerblue}{#1}}

\newacro{lvm}[LVM]{Latent Variable Model}
\newacro{gp}[GP]{Gaussian Process}
\newacro{gplvm}[GPLVM]{Gaussian Process Latent Variable Model}
\newacro{gphlvm}[GPHLVM]{Gaussian Process Hyperbolic Latent Variable Model}
\newacro{gpdm}[GPDM]{Gaussian Process Dynamical Model}
\newacro{gphdm}[GPHDM]{Gaussian Process Hyperbolic Dynamical Model}
\newacro{wgd}[WGD]{Wrapped Gaussian distribution}

\allowdisplaybreaks

\title{\LARGE \bf
Taxonomy-aware Dynamic Motion Generation on Hyperbolic Manifolds\vspace{-0.15cm}
}

\author{Luis Augenstein$^{*1,2}$, Noémie Jaquier$^{*3}$, Tamim Asfour$^{2}$, and Leonel Rozo$^{4}$% <-this % stops a space
\thanks{This work was partially supported by the Wallenberg AI, Autonomous Systems and Software Program (WASP) funded by the Knut and Alice Wallenberg Foundation and by the European Union’s Horizon Europe Framework Programme under grant agreement No 101070596 (euROBIN).}% <-this % stops a space
\thanks{$^{*}$Equal contribution. $^{1}$TNG Technology Consulting GmbH, Munich, Germany. $^{2}$Karlsruhe Institute of Technology, Karlsruhe, Germany. $^{3}$KTH Royal Institute of Technology, Stockholm, Sweden. $^{4}$Italian Institute of Artificial Intelligence (AI4I), Turin, Italy. Correspondence to: \href{mailto:jaquier@kth.se}{\textrm{jaquier@kth.se}}}
}
\begin{document}

\makeatletter
\let\@oldmaketitle\@maketitle% Store \@maketitle
\renewcommand{\@maketitle}{\@oldmaketitle% Update \@maketitle to insert...
	\vspace{-8ex}
}
\makeatother
\maketitle

\thispagestyle{empty}
\pagestyle{empty}

\begin{strip}
    \centering
    \captionsetup{type=figure}
	\begin{subfigure}[b]{.26\linewidth}
		\includegraphics[width=\linewidth]{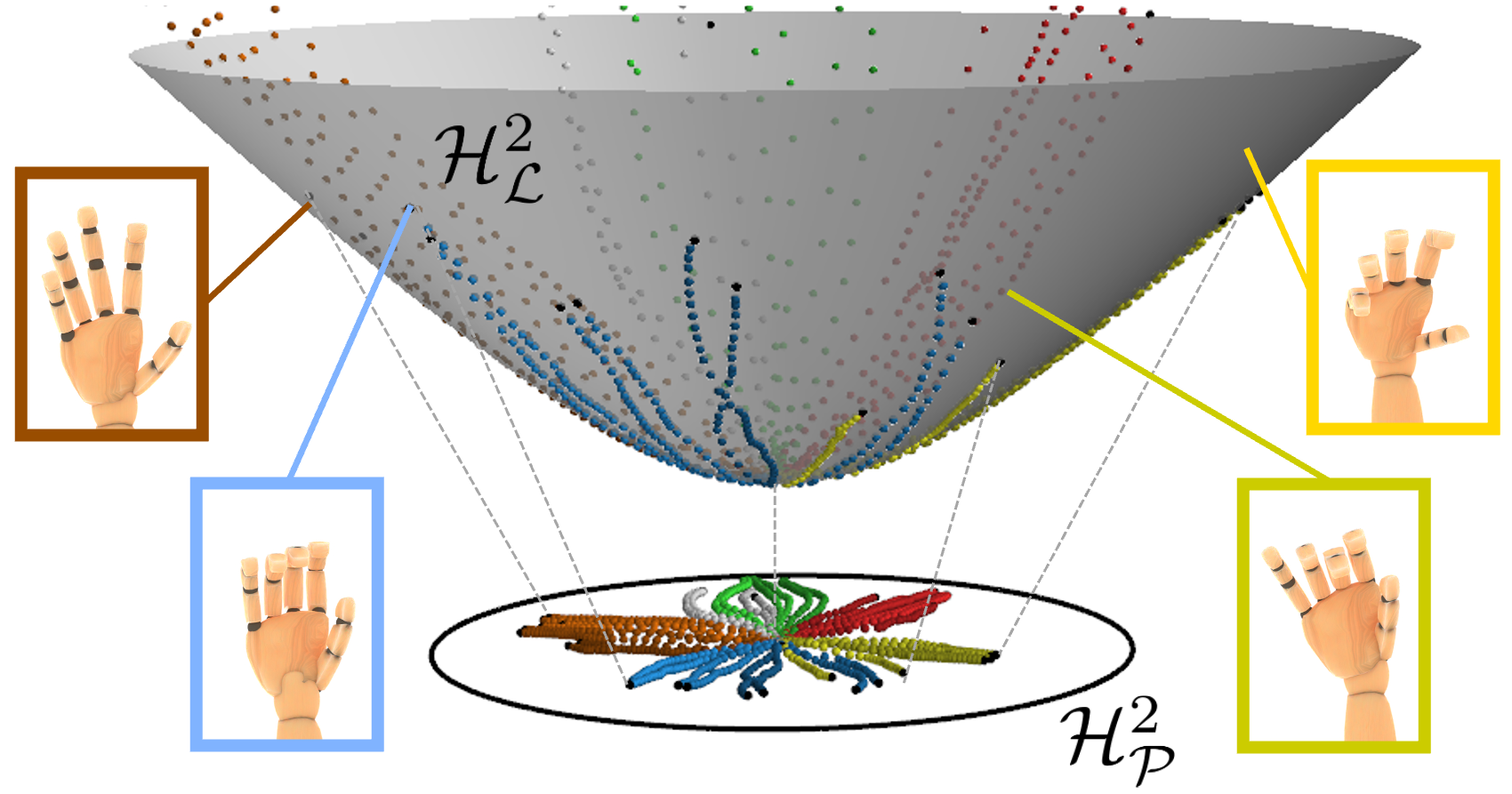}
	\end{subfigure}
 \hspace{0.3cm}
	\begin{subfigure}[b]{.61\linewidth}
		\includegraphics[width=\linewidth]{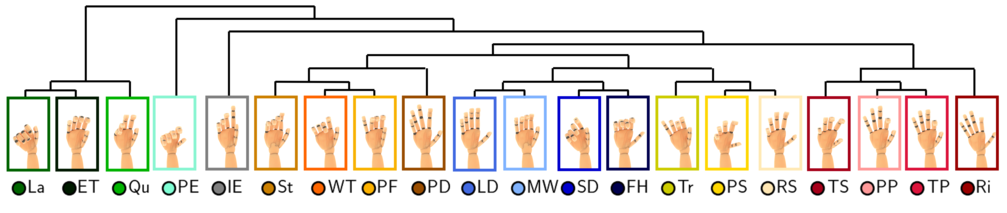}
	\end{subfigure}
    \vspace{-0.1cm}
	\caption{Illustration of latent trajectories learned by the proposed GPHDM on the Lorentz $\lorentz{2}$ and Poincaré $\poincare{2}$ models of the hyperbolic manifold. The GPHDM preserves the hand grasp taxonomy~\cite{Stival19:HumanGraspTaxonomy} (\emph{right}) and the temporal dynamics of the demonstrated hand motions. }
	\label{fig:teaser}
 \vspace{-0.3cm}
\end{strip}

%%%%%%%%%%%%%%%%%%%%%%%%%%%%%%%%%%%%%%%%%%%%%%%%%%%%%%%%%%%%%%%%%%%%%%%%%%%%%%%%
\begin{abstract}
Human-like motion generation for robots often draws inspiration from biomechanical studies, which categorize complex human motions into hierarchical taxonomies. While these taxonomies provide rich structural information about how movements relate to one another, this information is frequently overlooked in motion generation models, leading to a disconnect between the generated motions and their underlying hierarchical structure. This paper introduces the \ac{gphdm}, a novel approach that learns latent representations preserving both the hierarchical structure of motions and their temporal dynamics to ensure physical consistency.
Our model achieves this by extending the dynamics prior of the Gaussian Process Dynamical Model (GPDM) to the hyperbolic manifold and integrating it with taxonomy-aware inductive biases. Building on this geometry- and taxonomy-aware frameworks, we propose three novel mechanisms for generating motions that are both taxonomically-structured and physically-consistent: two probabilistic recursive approaches and a method based on pullback-metric geodesics. Experiments on generating realistic motion sequences on the hand grasping taxonomy show that the proposed GPHDM faithfully encodes the underlying taxonomy and temporal dynamics, and it generates novel physically-consistent trajectories.
\end{abstract}

%%%%%%%%%%%%%%%%%%%%%%%%%%%%%%%%%%%%%%%%%%%%%%%%%%%%%%%%%%%%%%%%%%%%%%%%%%%%%%%%

%%%%%%%%%%%%%%%%%%%%%%%%%%%%%%%%%%%%%%%%%%%%%%%%%%%%%%%%%%%%%%%%%%%%%%%%%%%%%%%%
\section{INTRODUCTION}
\label{sec:intro}
Designing robots with human-like capabilities is a long-standing goal in robotics, often drawing inspiration from biomechanics to achieve realistic and functional motions~\cite{Iida2016:BioinspiredRobots}. A critical aspect in this process is analyzing human movements, for which researchers often structure complex actions into hierarchical classifications known as \emph{taxonomies}. These taxonomies, which categorize hand postures~\cite{Feix16:GRASPtaxonomy}, full-body poses~\cite{Borras17:WholeBodyTaxonomy}, and manipulation primitives~\cite{Paulius20:ManipulationTaxonomy}, among others, carry a rich hierarchical structure that reflects the relationships among different movements. However, this crucial structural information is frequently ignored in motion generation literature. Recent works~\cite{Jaquier2024:GPHLVM,Augenstein25:Hyperbolic} have shown that leveraging this structural inductive bias into motion generation models may alleviate the demand of training data, lead to human-like movements, and ultimately generate new motions that comply with the hierarchical structure of a given taxonomy.

Early approaches to generating motions within these taxonomies showed promise but did not directly leverage their hierarchical nature. For example, Romero \etal~\cite{Romero10:SpatioTempGraspsGPLVM} used a \ac{gplvm}~\cite{Lawrence03:GPLVM}, where clusters corresponding to various grasp types of the GRASP taxonomy~\cite{Feix16:GRASPtaxonomy} emerged naturally, despite that the taxonomy's structure was not explicitly considered during training. They identified the potential for generating new motions via latent space interpolation as potential future work. Separately, Mandery \etal~\cite{Mandery16:LanguageWholeBody} abstracted the motion generation problem as a linguistic task, using probabilistic $n$-gram language models to reproduce whole-body motions based on the types of whole-body poses proposed in~\cite{Borras17:WholeBodyTaxonomy}. Discrete body poses were represented by words and sequenced using sentences. However, this discrete representation struggled to capture the continuous nature of movement and overlooked the entire hierarchical structure of the associated taxonomy.

To address this gap, recent work by Jaquier \etal~\cite{Jaquier2024:GPHLVM} introduced the \ac{gphlvm}, which explicitly accounts for the hierarchical structure of taxonomy data. Their key insight was to leverage \emph{hyperbolic geometry}~\cite{Ratcliffe19:HyperbolicManifold}, a natural fit for embedding tree-like structures~\cite{Nickel2017:Poincare,Reynolds93:Hyperboloid,Chami2020:TreesHyperbolic}, to create a continuous latent representation of the hierarchically-organized taxonomy data. In the \ac{gphlvm}, high-dimensional observations (e.g., joint angles of a human body or hand) belonging to the same taxonomy node were embedded closely together, forming distinct clusters in the hyperbolic space. Crucially, they demonstrated that hyperbolic geodesics between these clusters correctly traversed intermediate clusters, mirroring the parent--child relationships of the original taxonomy. These insights confirmed that the \ac{gphlvm} could successfully learn a latent space that preserves the hierarchical structure of complex motion data. 
However, a significant limitation remains: While the \ac{gphlvm} can generate novel motions by decoding latent geodesics back to the high-dimensional joint space, some of the resulting motions can be physically impractical. This issue arises because the \ac{gphlvm} is trained on data concentrated within the clusters (i.e., static poses) of the taxonomy, leaving data-sparse regions in-between. In these regions, the model lacks information about valid trajectories and thus its predictions revert to non-informative mean, failing to capture the underlying dynamics of target movements.
 
\textbf{This paper} tackles this challenge. First, we propose the Gaussian Process Hyperbolic Dynamical Model (GPHDM), which learns latent representations that preserve not only the hierarchical structure of motions but also their temporal dynamics to ensure physical consistency (see Fig.~\ref{fig:teaser}). To achieve this, we extend the dynamics model of the \ac{gpdm}~\cite{Wang08:GPDM} to the hyperbolic manifold and integrate it into the taxonomy-aware framework of \ac{gphlvm}. Second, we introduce three novel motion generation mechanisms that are both taxonomically-structured and physically-consistent, namely: two probabilistic recursive approaches, and a pullback-metric geodesics method. We test our approach on the hand grasping taxonomy~\cite{Stival19:HumanGraspTaxonomy}, showing that the proposed \ac{gphdm} successfully preserves both the hierarchical structure and temporal dynamics of the data, while allowing us to generate novel physically-consistent motions. The source code is available at \url{https://github.com/NoemieJaquier/hyperbolic-gplvms}.

\section{BACKGROUND}
\label{sec:background}
\textbf{Riemannian geometry:}
\label{sec:background-riemannian-geometry}
A Riemannian manifold $\manifold$ is a smooth manifold equipped with a Riemannian metric, i.e., a smoothly-varying inner product ${g_{\bm{x}}\!:\mathcal{T}_{\bm{x}}\manifold \! \times \! \mathcal{T}_{\bm{x}}\manifold \!\rightarrow \! \euclideanspace}$ over tangent spaces $\mathcal{T}_{\bm{x}}\manifold$~\cite{Lee18:RiemannManifold}. When considering coordinates, the Riemannian metric is represented in matrix form as $\innerprod{\bm{x}}{\bm{u}}{\bm{v}} \!=\! \bm{u}^\trsp \metric_{\bm{x}} \bm{v}$ with $\bm{u},\bm{v}\in\mathcal{T}_{\bm{x}}\manifold$. This metric defines the length of curves in $\manifold$, leading to the definition of geodesics, defined as locally length-minimizing curves.
To operate with Riemannian manifolds, one can leverage the Euclidean tangent spaces and the geodesics. 
The exponential map $\expmap{\bm{x}}{\bm{u}} \!=\! \bm{y}$ maps $\bm{u} \in \mathcal{T}_{\bm{x}}\manifold$ to a point $\bm{y}\in\manifold$, so that $\bm{y}$ lies on a geodesic starting at $\bm{x}$ in the direction $\bm{u}$, and such that the geodesic distance $\manifolddist{\bm{x}}{\bm{y}}$ equals the length of $\bm{u}$ given by $\norm{\bm{x}}{\bm{u}} \!=\! \sqrt{\innerprod{\bm{x}}{\bm{u}}{\bm{v}}}$. The inverse operation defines the logarithmic map $\logmap{\bm{x}}{\bm{y}} \!=\! \bm{u}$. Finally, the parallel transport $\Gamma_{\bm{x}\rightarrow \bm{y}} (\bm{u}) = \bm{v}$ moves a vector $\bm{u}\in\mathcal{T}_{\bm{x}}\manifold$ to $\mathcal{T}_{\bm{y}}\manifold$ while preserving the Riemannian inner product. \looseness-1

% Optimizing functions $f: \manifold \to \euclideanspace$ requires generalizing the definition of gradients. The Riemannian gradient $\grad_{\bm{x}}f$ of $f$ at $\bm{x}\in\manifold$ is the unique tangent vector in $\mathcal{T}_{\bm{x}} \manifold$ satisfying $\mathcal{D} _{\bm{u}} f(\bm{x})\!=\! \langle \grad_{\bm{x}} f, \bm{u} \rangle_{\bm{x}}$, with $\mathcal{D}_{\bm{u}} f(\bm{x})$ the directional derivative of $f$ along $\bm{u}\in\mathcal{T}_{\bm{x}}\manifold$~\cite[Chap. 3]{Boumal22:RiemannOpt}. The Riemannian Jacobian $\bm{J}^{\manifold}$ of ${f: \manifold \to \euclideanspace^{D}}$ is composed by the Riemannian gradients for each output dimension, i.e., 
% \begin{equation}
% \bm{J}^{\manifold} = [\grad_{\bm{x}}f_1 \:\ldots\: \grad_{\bm{x}}f_{D}]^\trsp .
% \label{eq:RiemannianJacobian}
% \end{equation}

\textbf{Hyperbolic Manifold:}
\label{sec:background-hyperbolic-manifold}
The hyperbolic manifold is the only Riemannian manifold with constant negative curvature~\cite{Ratcliffe19:HyperbolicManifold}. It is often represented by either the Poincaré model $\mathbb{H}^{D}_{\mathcal{P}}$~\cite{Poincare00:PoincareModel}, using local coordinates within the unit ball, or the Lorentz model $\lorentz{D}$~\cite{Jansen09:Hyperboloid,Reynolds93:Hyperboloid}, using Cartesian coordinates to represent the surface embedded in $\euclideanspace^{D+1}$ (see Fig.~\ref{fig:teaser}-\emph{left}). We mostly rely on the latter as it is numerically more stable. The Lorentz model is defined as ${\lorentz{D} = \{ \bm{x} \in \euclideanspace^{D+1} \mid \langle \bm{x}, \bm{x} \rangle_\mathcal{L} = -1, x_0 > 0 \} }$,
where $\langle \bm{x}, \bm{y} \rangle_\mathcal{L} = \bm{x}^\trsp \lorentzmetric \bm{y}$ is the Lorentzian inner product with metric $\lorentzmetric = \text{diag}(-1, 1, ..., 1)$. Under this model, the Riemannian operations of relevance for this work are,
\begin{align}
    d_{\lorentz{D}}(\bm{x}, \bm{z}) &= \text{arccosh}(-\langle \bm{x}, \bm{z} \rangle_{\mathcal{L}}) \, , \\
    \expmap{\bm{x}}{\bm{u}} &= \cosh(\norm{\bm{x}}{\bm{u}}) \bm{x} + \sinh(\norm{\bm{x}}{\bm{u}}) \dfrac{\bm{u}}{\norm{\bm{x}}{\bm{u}}} \, , \\
    \logmap{\bm{x}}{\bm{z}} &= \text{d}_{\lorentz{D}}(\bm{x}, \bm{z}) \dfrac{\bm{z} + \langle \bm{x}, \bm{z} \rangle_{\mathcal{L}} \bm{x}}{\sqrt{\langle \bm{x}, \bm{z} \rangle_{\mathcal{L}}^2-1}} \, , \\
    \Gamma_{\bm{x} \rightarrow \bm{z}}(\bm{u}) &= \bm{u} + \dfrac{\langle \bm{z}, \bm{u}\rangle_{\mathcal{L}}}{1 - \langle \bm{x}, \bm{z}\rangle_{\mathcal{L}}} (\bm{x} + \bm{z}) \\
    \proj_{\bm{x}} (\bm{w}) &= \bm{P}_{\bm{x}} \bm{w} = (\bm{I} + \bm{x}\bm{x}^\trsp\lorentzmetric)\bm{w} \, , \label{eq:proj_op}
\end{align}
where $\proj_{\bm{x}} (\bm{w})$ projects $\bm{w}\in\euclideanspace^{D+1}$ to $\tangentspacelorentz{\bm{x}}{D}$.

\textbf{Hyperbolic wrapped Gaussian distribution:}
Working with probabilistic models on Riemannian manifolds requires probability distributions that account for their geometry.
We employ the \ac{wgd}, which maps a normal distribution from a Euclidean tangent space onto a Riemannian manifold $\manifold$. This \emph{wrapping} is performed by a diffeomorphism $\psi$, which is often the exponential map $\expmap{\bm{\mu}}{\cdot}$, centered at the distribution's mean $\bm{\mu}$ and restricted to a domain where it is injective. Specifically, a \ac{wgd} is defined as the pushforward measure $\psi_{\#}(\mathcal{N})$ of a Gaussian distribution with density $\mathcal{N}$. The wrapping operation induces a wrapped density function through the change of variables, 
\begin{equation}
	\mathcal{N}_W = \mathcal{N} \circ \psi^{-1} \cdot |\det(\partial \psi^{-1})| .
	\label{Eq:ChangeVar_WGD}
\end{equation}   
Given the Lorentz model as the representation of a hyperbolic manifold, the hyperbolic \ac{wgd} corresponds to,
\begin{equation}
    \mathcal{N}_{\lorentz{D_x}} = \gaussiandist{\logmap{\bm{\mu}}{\bm{x}}}{\bm{0}}{\bm{\Sigma}} \cdot \left(\frac{\norm{\bm{\mu}}{\logmap{\bm{\mu}}{\bm{x}}}}{\sinh(\norm{\bm{\mu}}{\logmap{\bm{\mu}}{\bm{x}}})}  \right)^{D_x-1} .
    \label{Eq:HyperbolicWGD}
\end{equation}

\textbf{Gaussian Process Hyperbolic Latent Variable Model (GPHLVM):}
\label{sec:background-gphlvm}
A \ac{gplvm} defines a stochastic mapping from latent variables ${\bm{X} = [\bm{x}_1,  \ldots, \bm{x}_N
]^\top}$, $\bm{x}_n \!\in\! \euclideanspace^{D_x}$ to observations $\bm{Y} = [\bm{y}_1,  \ldots, \bm{y}_N
]^\top$, $\bm{y}_n\!\in\! \euclideanspace^{D_y}$ with $D_x \!<\! D_y$ via a non-linear transformation modeled by a \ac{gp}~\cite{Lawrence03:GPLVM}. 
 A \ac{gplvm} defines a generative model
 \begin{equation}
    p(\bm{Y} ; \bm{X}, \Theta) = \, \gaussiandist{\bm{Y}}{\bm{0}}{ k(\bm{X,\bm{X}}) + \sigma^2_{\bm{y}}\bm{I}} \, ,
    \label{eq:GPLVM}
\end{equation}
with prior $\bm{x}_n \sim \mathcal{N}(\bm{0},\bm{I})$, where $k:\euclideanspace^{D_x}\times \euclideanspace^{D_x} \to \euclideanspace$ is the GP kernel function that measures the similarity between pairs $(\bm{x}_n,\bm{x}_m)$, and $\Theta$ denotes the model hyperparameters, i.e., the kernel parameters and the noise variance $\sigma_{\bm{y}}^2$. 

In the hyperbolic setting, the latent variables $\bm{x}_n \in \mathbb{H}_{\mathcal{L}}^{D_x}$ lie in a hyperbolic space, so the generative mapping is defined via a \ac{gphlvm}~\cite{Jaquier2024:GPHLVM}. Therefore, the kernel function in~\eqref{eq:GPLVM} is replaced by a hyperbolic kernel $k^{\mathbb{H}_{\mathcal{L}}^{D_x}}:\mathbb{H}_{\mathcal{L}}^{D_x}\times \mathbb{H}_{\mathcal{L}}^{D_x} \to \euclideanspace$. Here, we employ hyperbolic kernels that build on the definition of the heat kernel on Riemannian manifolds \cite{GrigoryanNoguchi98:HyperbolicHeatKernel,Jaquier21:GaBOMatern,Jaquier2024:GPHLVM}, which accurately capture the non-Euclidean geometry of the hyperbolic space. 
Moreover, the latent variables follow a hyperbolic wrapped Gaussian prior
$\bm{x}_n \sim \mathcal{N}_{\mathbb{H}^{D_x}_{\mathcal{L}}}(\bm{\mu}_0, \alpha \bm{I})$, where $\bm{\mu}_0=(1, 0, \ldots, 0)^\trsp$ is the hyperbolic origin and $\alpha$ controls the spread of the latent variables. 
The \ac{gphlvm} latent variables and hyperparameters are inferred via MAP or variational inference similar as in the Euclidean case~\cite{Jaquier2024:GPHLVM}.\looseness=-1 

\textbf{Pullback metrics:}
An immersion ${f \colon \mathcal{X}\to \mathcal{Y}}$ from a latent space $\mathcal{X}$ to a Riemannian manifold $\mathcal{Y}$ equipped with a Riemannian metric $g_{\bm{y}}$ induces a pullback metric $g_{\bm{x}}^{\text{P}}$ on $\mathcal{X}$ which, for $\bm{x}\in \mathcal{X}$ and $\bm{u}, \bm{v} \in \mathcal{T}_{\bm{x}}\mathcal{X}$, is given by~\cite[Chap.2]{Lee18:RiemannManifold}, 
\begin{equation}
    g^{\text{P}}_{\bm{x}}(\bm{u}, \bm{v}) = g_{f(\bm{x})}\big(d f_{\bm{x}}(\bm{u}), d f_{\bm{x}}(\bm{v})\big). 
    \label{eq:pullback}
\end{equation}
In coordinates, the pullback metric is given in matrix form by $\pullbackmetric_{\bm{x}} = {\bm{J}^{\mathcal{X}}}^\trsp \metric_{\bm{y}} \bm{J}^{\mathcal{X}}$, 
where $\bm{J}^{\mathcal{X}}$ is the Jacobian of $f$ at $\bm{x}$ with domain restricted to $\mathcal{T}_{\bm{x}}\mathcal{X}$. Intuitively, $g^{\text{P}}_{\bm{x}}$ evaluates on tangent vectors of $\mathcal{T}_{\bm{x}}\mathcal{X}$ by moving them to $\mathcal{T}_{f(\bm{x})}\mathcal{Y}$ to compute their inner product. For an immersion $f:\mathcal{X}\to\euclideanspace^{D_y}$ with Euclidean co-domain, i.e., $\metric_{\bm{y}} = \bm{I}$, the pullback metric is defined as $\pullbackmetric_{\bm{x}} = \bm{J}^\trsp \!\bm{J}$ with the Jacobian ${\bm{J}=[\frac{\partial f_1}{\partial \bm{x}} \:\ldots\: \frac{\partial f_{D_y}}{\partial \bm{x}}]^\trsp\in \euclideanspace^{D_y \times D_x}}$. 

%%%%%%%%%%%%%%%%%%%%%%%%%%%%%%%%%%%%%%%%%%%%%%%%%%%%%%%%%%%%%%%%%%%%%%%%%%%%%%%%
\section{GAUSSIAN PROCESS HYPERBOLIC DYNAMICAL MODEL}
\label{chap:content}
The standard \ac{gplvm}, introduced in Sec.~\ref{sec:background}, does not account for the temporal structure of the data. To address this, the \ac{gpdm}~\cite{Wang08:GPDM} extends the \ac{gplvm} by adding a dynamics prior $p(\bm{X})$ over the latent variables. This prior, derived under a first-order Markov assumption, incentivizes the latent points to form smooth trajectories. A \ac{gpdm} often assumes a linear model $f_{\bm{A}}(\bm{x}_t) = \sum_{i=1}^{N_\phi} \bm{a}_i \phi_i(\bm{x}_t) = \bm{A}^\trsp \bm{\phi}_t$ for the dynamics. However, we introduce an alternative formulation that models only the offset required to transition from $\bm{x}_t$ to $\bm{x}_{t+1}$ as, 
\begin{equation}
\label{eq:EuclideanMarkovDynamics}
f_{\bm{A}}(\bm{x}_t) = \bm{x}_t + \sum_{i=1}^{N_\phi} \bm{a}_i \phi_i(\bm{x}_t) = \bm{x}_t + \bm{A}^\trsp \bm{\phi}_t .    
\end{equation}
This can be equivalently written as $\tilde{\bm{x}}_{t+1} := \bm{x}_{t+1} - \bm{x}_t = \bm{A}^\trsp \bm{\phi}_t + \bm{\epsilon}_t$, with   $\bm{\epsilon}_t \sim \mathcal{N}(\bm{0}, \bm{\Sigma}_x)$. While this modification may seem unnecessary in the Euclidean case, it provides a clearer connection for comparing the classical \ac{gpdm} with our method, which places a hyperbolic dynamics prior on the latent space of the \ac{gphlvm}, as explained shortly. 

\subsection{Hyperbolic Tangent Vectors in Local Coordinates}
The operations introduced for the Lorentz model of the hyperbolic manifold in Sec.~\ref{sec:background} represent tangent vectors as $(D_x\!+\!1)$-dimensional elements embedded in the ambient space. However, the intrinsic dimension of the hyperbolic manifold and its tangent space is $D_x$, which leads to degenerated covariance matrices $\bm{\Sigma}\in\tangentspacelorentz{\bm{x}}{D_x}$, e.g., in the hyperbolic \ac{wgd}~\eqref{Eq:HyperbolicWGD}, since the eigenvalue along the eigenvector orthogonal to $\tangentspacelorentz{\bm{x}}{D_x}$ is always $0$. In this paper, we apply a local change of coordinates to represent tangent vectors and covariances intrinsically as $D_x$-dimensional elements, thus avoiding issues related to degenerated covariances in the \ac{wgd} akin to~\cite{Rozo2025:Riemann2}.

We first define a canonical set of $D_x$ basis vectors ${\bm{V}_{\bm{\mu}_0} = \left(\begin{matrix}
\bm{e}_2 & \ldots & \bm{e}_{D+1}
\end{matrix}\right) = \left(\begin{matrix} 
\bm{0} &
\bm{I}_{D_x}
\end{matrix}\right)^\top \in \euclideanspace^{(D_x+1) \times D_x}}$ in the tangent space $\tangentspacelorentz{\bm{\mu}_0}{D_x}$ at the origin $\bm{\mu}_0\in\lorentz{D_x}$. A set of basis vectors $\bm{V}_{\bm{x}}$ at any point $\bm{x}\in\lorentz{D_x}$ can then be obtained by parallel transporting the canonical basis vectors as,
\begin{equation}
\bm{V}_{\bm{x}} = \left(\begin{matrix}
\Gamma_{\bm{\mu}_0 \rightarrow \bm{x}}(\bm{e_2}) & \ldots & \Gamma_{\bm{\mu}_0\rightarrow \bm{x}}(\bm{e}_{D+1})
\end{matrix}\right).
\end{equation}
Note that $\bm{V}_{\bm{x}}$ consists of orthonormal basis vectors on $\tangentspacelorentz{\bm{x}}{D_x}$, i.e., $\bm{V}_{\bm{x}}^\top \lorentzmetric \bm{V}_{\bm{x}} \!= \! \bm{I}_{D_x}$ and $\bm{P}_{\bm{x}} = \bm{V}_{\bm{x}} \bm{V}_{\bm{x}}^\top\lorentzmetric$. 
We obtain tangent space vectors $\bm{u}$ and matrices $\bm{\Sigma}\in\tangentspacelorentz{\bm{x}}{D_x}$ in \localcord{local coordinates} using $\bm{V}_{\bm{x}}$ as $\localcord{\tilde{\bm{u}}} = \bm{V}_{\bm{x}}^\top \lorentzmetric \bm{p}$ and $\localcord{\tilde{\bm{\Sigma}}}=\bm{V}_{\bm{x}}^\trsp \lorentzmetric \bm{\Sigma} \lorentzmetric\bm{V}_{\bm{x}}$.
Tangent space elements in ambient space coordinates are obtained via the inverse operations $\bm{u} = \bm{V}_{\bm{x}} \localcord{\tilde{\bm{u}}}$ and $\bm{\Sigma}=\bm{V}_{\bm{x}}^\trsp \localcord{\tilde{\bm{\Sigma}}} \bm{V}_{\bm{x}}$.\looseness=-1  

\subsection{Hyperbolic Dynamics Prior}
\label{subsec:HypeDynamicsPrior}
To derive the \ac{gphdm}, we first generalize the dynamics prior of the \ac{gpdm} to the hyperbolic manifold. 
To do so, we define a first-order hyperbolic dynamics model by extending the Markov assumption to the hyperbolic manifold as,
\begin{equation}
    \bm{x}_{t+1} = \expmap{f_{\bm{A}}(\bm{x}_t)}{\bm{V}_{f_{\bm{A}}(\bm{x}_t)} \localcord{\tilde{\bm{\epsilon}}_t}} \;\; \text{with} \;\; \localcord{\tilde{\bm{\epsilon}}_t} \sim \mathcal{N}(\localcord{\bm{0}}, \localcord{\tilde{\bm{\Sigma}}_{\bm{x}}}),
    \label{eq:HyperbolicMarkovDynamics}
\end{equation}
where we define the noise $\tilde{\bm{\epsilon}}_t \in \euclideanspace^{D_x}$ in \localcord{local coordinates} and transform it onto the tangent space $\tangentspacelorentz{f_{\bm{A}}(\bm{x}_t)}{D_x}$ via the basis-vector matrix $\bm{V}_{f_{\bm{A}}(\bm{x}_t)} \in \euclideanspace^{(D_x+1) \times D_x}$. 
Similar to the \ac{gpdm} in~\eqref{eq:EuclideanMarkovDynamics}, we define $f_{\bm{A}}$ via $N_{\phi}$ nonlinear basis functions as, 
\begin{equation}
   f_{\bm{A}}(\bm{x}_t) = \expmap{\bm{x}_t}{\bm{V}_{\bm{x}_t} \localcord{\bm{A}^\top \bm{\phi}_t}},
\end{equation}
where ${\bm{\phi}_t = \bm{\phi}(\bm{x}_t) = \begin{bmatrix}
    \phi_1(\bm{x}_t) & ... & \phi_{N_{\phi}}(\bm{x}_t)
\end{bmatrix}^\top \in \euclideanspace^{N_{\phi}}}$ is a vector of basis functions and $\bm{A}\!\in\!\euclideanspace^{N_{\phi} \times D_x}$ is a weight matrix. 
Notice that we again perform a change of basis to obtain the Lorentz representation $\bm{V}_{\bm{x}_t} \localcord{\bm{A}^\top \bm{\phi}_t} \in \tangentspacelorentz{\bm{x}_t}{D_x}$ from the vector $\localcord{\bm{A}^\top \bm{\phi}_t} \!\in\! \euclideanspace^{D_x}$ represented in \localcord{local coordinates}.
In the Euclidean case, we recover the first-order Markov dynamics~\eqref{eq:EuclideanMarkovDynamics}. \looseness-1

The first-order hyperbolic Markov dynamics~\eqref{eq:HyperbolicMarkovDynamics} lead to the following transition probability,
\begin{align}
    \label{eq:hyperbolic-markov}
    p(\bm{x}_{t+1} \mid \bm{x}_t, \bm{A}) &= \hypenormallatent{\bm{x}_{t+1}}{f_{\bm{A}}(\bm{x}_t)}{ \bm{\Sigma}_{\bm{x}_t}}  \\
    &= \gaussiandist{\logmap{\bm{x}_t}{\bm{x}_{t+1}}} {\bm{V}_{\bm{x}_t} \localcord{\bm{A}^\top \bm{\phi}_t}}{\bm{\Sigma}_{\bm{x}_t}} \,r_t 
    \nonumber \\
    \label{eq:hyperbolic-markov-local}
    &= \gaussiandist{\localcord{\tilde{\bm{x}}_{t+1}}}{\localcord{\bm{A}^\top \bm{\phi}_t}}{\localcord{\tilde{\bm{\Sigma}}_{\bm{x}}}} \, r_t ,
\end{align}
where the \ac{wgd} $\mathcal{N}_{\lorentz{D_x}}$ is defined as in~\eqref{Eq:ChangeVar_WGD} with a pushforward $\psi=\expmap{\bm{x}_t}{\cdot}$, resulting into the change of volume ${r_t=|\det \partial \psi^{-1}|=\left(\frac{\norm{\bm{x}_t}{\logmap{\bm{x}_t}{\bm{x}_{t+1}}}}{\sinh(\norm{\bm{x}_t}{\logmap{\bm{x}_t}{\bm{x}_{t+1}}})}  \right)^{D_x-1}}$, and noise covariance matrix $\bm{\Sigma}_{\bm{x}_t} = \bm{V}_{\bm{x}_t} \localcord{\tilde{\bm{\Sigma}}_{\bm{x}}} \bm{V}_{\bm{x}_t}^\top \in \tangentspacelorentz{\bm{x}_t}{D_x}$ obtained from the common local covariance $\localcord{\tilde{\bm{\Sigma}}_{\bm{x}}}\in\euclideanspace^{D_x \times D_x}$. Equation~\eqref{eq:hyperbolic-markov-local} represents the Gaussian distribution in local coordinates with  $\localcord{\tilde{\bm{x}}_{t+1}} = \bm{V}_{\bm{x}_t}^\top \lorentzmetric \logmap{\bm{x}_t}{\bm{x}_{t+1}} \in \euclideanspace^{D_x}$.

We derive the hyperbolic dynamics prior $p(\bm{X})$ by marginalizing out the parameters $\bm{A}$, similar to the Euclidean case. For a trajectory $\bm{X}$ of $N$ latent variables $\bm{x}_1, ..., \bm{x}_N \in \lorentz{D_x}$ the hyperbolic dynamics prior is given as,
\begin{equation}
    p(\bm{X}) = \int p(\bm{X} \mid \bm{A}) \, p(\bm{A}) \, d \bm{A}.
\end{equation}
Incorporating the Markov probability transition~\eqref{eq:hyperbolic-markov-local} leads to,
\begin{align*}
    p(\bm{X}) &= p(\bm{x}_1) \int \prod_{t=1}^{N-1} p(\bm{x}_{t+1} \mid \bm{x}_{t}, \bm{A}) \, p(\bm{A}) \, \text{d} \bm{A}  \\
    &\stackrel{\eqref{eq:hyperbolic-markov-local}}{=} p(\bm{x}_1) \int \prod_{t=1}^{N-1} \gaussiandist{\localcord{\tilde{\bm{x}}_{t+1}} }{\localcord{ \bm{A}^\top \bm{\phi}_{t}}}{ \localcord{\tilde{\bm{\Sigma}}_{\bm{x}}}} \,r_t \, p(\bm{A}) \, \text{d} \bm{A}.
\end{align*}
We leverage the properties of the Euclidean Gaussian distributions to reduce the expression as follows,
\small
\begin{align*}
    p(\bm{X}) &= p(\bm{x}_1) \prod_{d=1}^{D_x} \int \prod_{t=1}^{N-1} p(\localcord{\tilde{x}_{t+1,d}} ; \localcord{\bm{\phi}_{t}^\top \bm{A}_d}, \localcord{\sigma_{x,d}^2}) \,r_t\, p(\bm{A}_d) \, \text{d} \bm{A}_d 
\end{align*}
\normalsize
\begin{align*}
    \quad &= p(\bm{x}_1) \prod_{t=1}^{N-1} r_t \prod_{d=1}^{D_x} \gaussiandist{\localcord{\tilde{\bm{X}}_{2:N, d}}}{ \localcord{\bm{0}}}{ \localcord{\bm{\Phi} \bm{\Phi}^\top + \sigma_{x,d}^2 \bm{I}_{N-1}}} \\
    &= p(\bm{x}_1) \prod_{t=1}^{N-1} r_t \prod_{d=1}^{D_x} \gaussiandist{\localcord{\tilde{\bm{X}}_{2:N, d}} }{ \localcord{\bm{0}}}{\localcord{\bm{K}_{\bm{X},d} + \sigma_{x,d}^2 \bm{I}_{N-1}}} \\
    &= p(\bm{x}_1) \prod_{t=0}^{N-1} r_t \, \gaussiandist{\localcord{\tilde{\bm{X}}_{2:N}} }{ \localcord{\bm{0}}}{ \localcord{\bm{K}_{\bm{X}} + \tilde{\bm{\Sigma}}_{\bm{x}}}},
\end{align*}
where $\bm{A}_d$ is the $d$-th column of $\bm{A}$, $\localcord{\tilde{\bm{X}}_{2:N}} = [\localcord{\tilde{\bm{x}}_2}, \ldots,\localcord{\tilde{\bm{x}}_N} ]^\top$, and  $\localcord{\bm{K}_{\bm{X}}}=\bm{\Phi}\bm{\Phi}^\top$ is the $(N\!-\!1)D_x\times(N\!-\!1)D_x$ kernel matrix constructed from the $(N-1)$ basis functions associated to $\bm{x}_1,\ldots,\bm{x}_{N-1}$. Also, we used $p(\bm{A}_d)=\gaussiandist{\bm{A}_d}{\bm{0}}{\bm{I}_{N_{\phi}}}$ and
\small
\begin{equation*}
    \prod_{t=1}^{N-1} p(\localcord{\tilde{x}_{t+1,d}} ; \localcord{\bm{\phi}_{t}^\top \bm{A}_d}, \localcord{\sigma_{x,d}^2}) = \gaussiandist{\localcord{\tilde{\bm{X}}_{2:N, d}} }{ \localcord{\bm{\Phi} \bm{A}_d }}{ \localcord{\sigma_{x,d}^2 \bm{I}_{N-1}}}.
\end{equation*}
\normalsize
Finally, we obtain the hyperbolic dynamics prior as,
\begin{align}
    \label{eq:HyperbolicDynamicsPrior}
    p(\bm{X})= \,&\hypenormallatent{\bm{x}_1 }{ \bm{\mu}_0}{ \bm{V}_{\bm{\mu}_0} \bm{V}_{\bm{\mu}_0}^\top}  \\
    &\hypenormallatent{\bm{X}_{2:N} }{ \bm{X}_{1:N-1}}{ \bm{V}_{\bm{X}}(\localcord{\bm{K}_{\bm{X}}} + \localcord{\tilde{\bm{\Sigma}}_{\bm{x}}}) \bm{V}_{\bm{X}}^\top}, \nonumber
\end{align}
where $p(\bm{x}_1)$ was set as an isotropic hyperbolic \ac{wgd}, and ${\bm{V}_{\bm{X}} = \operatorname{blockdiag}(\bm{V}_{\bm{x}_1}, ..., \bm{V}_{\bm{x}_{N-1}}) \in \euclideanspace^{(N-1)(D_x+1) \times (N-1) D_x}}$. 

The kernel $k_{\bm{X}}$, used as $\bm{K}_{\bm{X}}=k_{\bm{X}}(\bm{X}_{1:N-1}, \bm{X}_{1:N-1})$ in~\eqref{eq:HyperbolicDynamicsPrior}, is key for the hyperbolic dynamics prior as it provides a notion of similarity between consecutive hyperbolic latent variables. Therefore, we must employ hyperbolic kernels that accurately capture the geometry of the hyperbolic space. Here, we leverage the hyperbolic squared exponential (SE) kernels $k^{\lorentz{D_x}}$~\cite{GrigoryanNoguchi98:HyperbolicHeatKernel,Jaquier21:GaBOMatern,Jaquier2024:GPHLVM}, and define $k_{\bm{X}}=\bm{I}_{D_x}\otimes k^{\lorentz{D_x}}$ as a multivariate kernel with shared values across output dimensions with $\otimes$ denoting the Kronecker product. The hyperbolic SE kernels in dimensions $2$ and $3$ are defined as,
\begin{align}
\label{eq:hyp_heat2}
k^{\lorentz{2}}(\bm{x}, \bm{z})
&=
\frac{\sigma^2}{C_{\infty}}
\int_{\rho}^{\infty}
\frac{s e^{-s^2/(2 \kappa^2)}}{(\cosh(s) - \cosh(\rho))^{1/2}} \mathrm{d} s,
\\
\label{eq:hyp_heat3}
k^{\lorentz{3}}(\bm{x}, \bm{z})
&=
\frac{\sigma^2}{C_{\infty}}
\frac{\rho}{\sinh{\rho}} e^{-\rho^2/(2 \kappa^2)},
\end{align}
where $\rho = \hypedist{\bm{x}}{\bm{z}}$ is the geodesic distance between $\bm{x}, \bm{z} \in \lorentz{D_x}$,  $\kappa$ and $\sigma^2$ are the kernel lengthscale and variance, and $C_{\infty}$ is a normalizing constant. 
To the best of our knowledge, no closed form expression for $\lorentz{2}$ is known, therefore we use the Monte-Carlo approximation introduced in~\cite{Jaquier2024:GPHLVM}. Notice that higher dimensional hyperbolic SE kernels are expressed as derivatives of the kernels~\eqref{eq:hyp_heat2}-\eqref{eq:hyp_heat3}~\cite{GrigoryanNoguchi98:HyperbolicHeatKernel}.

\subsection{The GPHDM}
Our \ac{gphdm} extends the \ac{gpdm} to hyperbolic spaces by combining the GPHLVM with the hyperbolic dynamics prior~\eqref{eq:HyperbolicDynamicsPrior}. A \ac{gphdm} defines a generative model from low-dimensional latent variables ${\bm{X} = [\bm{x}_1,  \ldots, \bm{x}_N
]^\top}$, $\bm{x}_t\in\lorentz{D_x}$ to high-dimensional trajectories $\bm{Y} = [\bm{y}_1,  \ldots, \bm{y}_N
]^\top$, ${\bm{y}_t\in\euclideanspace^{D_y}}$, and it is formally described as, 
\begin{align}
    &p(\bm{Y} ; \bm{X}, \Theta) = \, \gaussiandist{\bm{Y}}{\bm{0}}{\bm{K}_{\bm{Y}} + \bm{\Sigma}_{\bm{y}}} \, , \\
    &p(\bm{X} ; \Theta) = \,\hypenormallatent{\bm{x}_1 }{ \bm{\mu}_0}{ \bm{V}_{\bm{\mu}_0} \bm{V}_{\bm{\mu}_0}^\top }  \\
    &\qquad \qquad \,\,\,\,\hypenormallatent{\bm{X}_{2:N} }{ \bm{X}_{1:N-1}}{ \bm{V}_{\bm{X}}(\localcord{\bm{K}_{\bm{X}}} + \localcord{\tilde{\bm{\Sigma}}_{\bm{x}}}) \bm{V}_{\bm{X}}^\top}, \nonumber
\end{align}
where $p(\bm{Y} ; \bm{X}, \Theta)$ is the model likelihood, $p(\bm{X} ; \Theta)$ is the hyperbolic dynamics prior~\eqref{eq:HyperbolicDynamicsPrior} from Sec.~\ref{subsec:HypeDynamicsPrior}, and $\Theta$ denotes the model hyperparameters, i.e., the parameters of the hyperbolic kernels $k_{\bm{Y}}$ and $k_{\bm{X}}$, and the noise variances $\bm{\Sigma_{\bm{y}}}$ and $\tilde{\bm{\Sigma}}_{\bm{x}}$.\looseness-1

As in the Euclidean case, training a \ac{gphdm} consists of optimizing the latent variables $\bm{x}_t\in\lorentz{D_x}$ along with the parameters $\Theta$ by maximizing the log posterior of the model,
\begin{equation}
    \label{eq:gphdm-loss}
    \bm{X}, \Theta = \argmax_{\bm{X},\Theta} \: \beta_1 \log p(\bm{Y} ; \bm{X}, \Theta) + \beta_2 \log p(\bm{X} ; \Theta),
\end{equation}
where $\beta_1,\beta_2$ balance the trade-off between the likelihood and dynamic prior. Analogous to its Euclidean counterpart, the hyperbolic dynamics prior encourages consecutive latent points $\bm{x}_t$ and $\bm{x}_{t+1}$, to be close in the hyperbolic latent space, therefore promoting smooth latent trajectories. 
Optimizing the \ac{gphdm} requires accounting for the hyperbolic geometry of the latent variables. To this end, we employ Riemannian optimization methods~\cite{Boumal22:RiemannOpt}. In this paper, we use Riemannian Adam~\cite{Becigneul19:RiemannianAdaptiveOpt} implemented in Geoopt~\cite{Kochurov20:geoopt} to optimize~\eqref{eq:gphdm-loss}.\looseness-1

Finally, back constraints, introduced for the GPLVM in~\cite{Lawrence06:BackConstrGPLVM}, define the latent variables as a function of the observations via an encoder function $x_{t,d} = g_d (\bm{Y}; w_d)$ with parameters $\{w_d\}_{d=1}^{D_x}$. This method facilitates the post-training incorporation of new observations into the latent space while preserving local similarities among the resulting embeddings. To ensure that latent variables lie on the hyperbolic manifold, we use a similar back-constraints mapping as in~\cite{Jaquier2024:GPHLVM}, namely,
\begin{equation*}
    \bm{x}_{t} = \expmap{\bm{\mu}_0}{\check{\bm{x}}_t} \text{ with } \check{x}_{t,d} = \sum_{m=1}^N w_{d,m} k^{\mathbb{R}^{D_y}}(\bm{y}_t,\bm{y}_m).
\label{eq:backconstraints}
\end{equation*}
In this case, the loss~\eqref{eq:gphdm-loss} is optimized for $\{w_d\}_{d=1}^{D_x}$ and $\Theta$.

\subsection{Incorporating Taxonomy Knowledge}
Our \ac{gphdm} allows us to learn smooth hyperbolic latent trajectories of high-dimensional motions. When these observations are associated with a robotics taxonomy, the latent variables must preserve the taxonomy's graph structure. This means the hyperbolic distance between pairs of latent variables should match their corresponding graph distances. We enforce this property by introducing the graph-distance information as inductive bias during the \ac{gphdm} training.

As shown in~\cite{Urtasun08:TopologicalGPLVM}, the \ac{gplvm} latent space can be modified by adding priors of the form $p(\bm{X})\propto e^{-\phi(\bm{X})/\sigma^2_{\phi}}$, where $\phi(\bm{X})$ is a function to minimize. This is equivalent to augmenting the \ac{gplvm} loss with a regularization term $-\beta_3\phi(\bm{X})$. As in~\cite{Jaquier2024:GPHLVM}, we augment the \ac{gphdm} loss with such a regularizer aiming at preserving the distances of the taxonomy graph. 
We define $\phi(\bm{X})$ as the stress loss,
\begin{equation}
\ell_{\text{stress}}(\bm{X}) = \sum_{i<j}
\big( \graphdist{c_i}{ c_j} - \hypedist{\bm{x}_i}{\bm{x}_j} \big)^2 ,
\label{eq:stressLoss}
\end{equation}
where $c_i$ denotes the taxonomy node to which the observation $\bm{y}_i$ belongs, and $d_{\mathbb{G}}$, $d_{\lorentz{D_x}}$ are the taxonomy graph distance and the geodesic distance on $\lorentz{D_x}$, respectively. 
The stress loss~\eqref{eq:stressLoss} aims at preserving all the taxonomy graph distances in the latent space $\lorentz{D_x}$. 
However, for trajectories that transition between taxonomy classes, the intermediate trajectory points are not assigned to any taxonomy node. In such cases, we adapt the computation by applying the stress loss~\eqref{eq:stressLoss} only to the start and end points of each trajectory, which are anchored to specific nodes. The hyperbolic dynamics prior then ensures that intermediate trajectory points form a smooth path between these points in the latent space $\lorentz{D_x}$, thereby preserving the taxonomy graph globally.

\section{DYNAMIC MOTION GENERATION}
\label{sec:generating-new-hyperbolic-latent-trajectories}
The generation of new high-dimensional motions via the GPHDM requires the creation of novel latent trajectories $\bm{X}^*$ in $\lorentz{D_x}$. Here we introduce three strategies to accomplish this on the hyperbolic manifold. First, we adapt the mean prediction and conditional optimization for recursive trajectory generation of the \ac{gpdm}~\cite{Wang08:GPDM} to the hyperbolic setting. Then, we propose a new approach that generates trajectories as geodesics on the pullback metric of the learned model.

\subsection{Recursive motion generation}
\label{subsec:recursive_motion}
In the Euclidean case, the mean prediction method constructs a latent trajectory sequentially. The prediction for the next step $\bm{x}_{t+1}^*$ is the mean of the conditional distribution at the current step $\bm{x}_t^*$. This is straightforward as the conditional of the Gaussian distribution is also a Gaussian, making the mean analytically available.
However, this convenient property does not hold for the hyperbolic \ac{wgd} as the conditional of a \ac{wgd} is not necessarily a \ac{wgd}~\cite{Mallasto18:wrapped}. Consequently, its mean is analytically intractable. To overcome this, we leverage a parallel from the Euclidean case: for a Gaussian distribution, the mean and the Maximum Likelihood Estimate (MLE) are equivalent. We therefore propose to find the MLE of the hyperbolic conditional distribution to determine the next step in the trajectory as follows,
\begin{align}
\label{eq:hyperbolic-mean-prediction}
&\bm{x}_{t+1}^* = \argmax_{\bm{x}_{t+1}^*} \:
\gaussiandist{\localcord{\tilde{\bm{x}}_{t+1}^*}} {\localcord{\tilde{\bm{m}}^*}}{\localcord{\tilde{\bm{\Sigma}}^*}} \: r_t, \quad \text{with}\\
&\quad \localcord{\tilde{\bm{x}}_{t+1}^*} =  \bm{V}_{\bm{x}_t^*}^\top  \lorentzmetric \logmap{\bm{x}_t^*}{\bm{x}_{t+1}^*} \nonumber \\
&\quad \localcord{\tilde{\bm{m}}^*} = \, \localcord{\bm{L}^*} \bm{V}_{\bm{X}}^\top  \lorentzmetric \logmap{\bm{X}_{1:N-1}}{\bm{X}_{2:N}}, \nonumber \\
&\quad \localcord{\tilde{\bm{\Sigma}}^*} = \localcord{k_{\bm{X}}}(\bm{x}_t^*, \bm{x}_t^*) - \localcord{\bm{L}^*} \localcord{k_{\bm{X}}}(\bm{X}_{1:N-1}, \bm{x}^*_t), \nonumber \\
&\quad \localcord{\bm{L}^*} = \localcord{k_{\bm{X}}}(\bm{x}^*_t, \bm{X}_{1:N-1}) (\localcord{\bm{K}_{\bm{X}}} + \localcord{\tilde{\bm{\Sigma}}_{\bm{x}}})^{-1}, \nonumber
\end{align}
where $r_t$ is the change of volume defined in Sec.~\ref{subsec:HypeDynamicsPrior}.
We solve the optimization problem in~\eqref{eq:hyperbolic-mean-prediction} using Riemannian Adam~\cite{Becigneul19:RiemannianAdaptiveOpt}. For quick convergence, we initialize the problem at $\bm{x}_{t+1}^* = \expmap{\bm{x}_t^*}{\bm{V}_{\bm{x}_t^*} \localcord{\tilde{\bm{m}}^*}}$, which maximizes the Euclidean Gaussian component of the objective~\eqref{eq:hyperbolic-mean-prediction}. The final solution $\bm{x}_{t+1}^*$ is usually located in the vicinity of this initial point.  

A key limitation of the mean prediction approach, analogous to the Euclidean case, is its inability to specify a desired goal point for the new latent trajectory. To address this, we extend the conditional optimization approach of~\cite{Wang08:GPDM} to the hyperbolic manifold. This method allows us to specify start, goal, and intermediate points, from which it interpolates the remaining trajectory segments according to learned dynamics prior. This approach optimizes the full conditional distribution, mirroring the objective defined for the Euclidean case,
\begin{align}
    \label{eq:hyperbolic-conditional-prediction}
    \bm{X}^* = &\argmax_{\bm{X}^*} \: p(\bm{Y}^* \mid \bm{X}^*, \bm{Y}, \bm{X}, \Theta) \, p(\bm{X}^* \mid \bm{X}, \Theta)  \\
    \nonumber
    &\quad \text{s.t.} \quad \bm{Y}^* = \bm{k}_{\bm{Y}}(\bm{X}^*,\bm{X})(\bm{K}_{\bm{Y}} + \bm{\Sigma}_{\bm{Y}})^{-1} \bm{Y} \, .
\end{align}
The main difference is that the conditional dynamics prior corresponds to its hyperbolic version, i.e.,
\begin{align}
&p(\bm{X}^* ; \bm{X}) = \gaussiandist{\localcord{\tilde{\bm{X}}_{2:M}^* }}{ \localcord{\tilde{\bm{m}}^*}}{ \localcord{\tilde{\bm{\Sigma}}^*}} \: \prod_{t=1}^{M-1} r_t , \quad \text{with}\\
& \quad \localcord{\tilde{\bm{X}}_{2:M}^*} = \bm{V}_{\bm{X}^*}^\top \lorentzmetric \logmap{\bm{X}_{1:M-1}^*}{\bm{X}_{2:M}^*}, \nonumber \\
& \quad \localcord{\tilde{\bm{m}}^*} = \localcord{\bm{L}^*} \bm{V}_{\bm{X}}^\top \lorentzmetric \logmap{\bm{X}_{1:N-1}}{\bm{X}_{2:N}}, \nonumber \\
&\quad\localcord{\tilde{\bm{\Sigma}}^*} = \, \localcord{\bm{K}_{\bm{X}}^*} - \localcord{\bm{L}^*} \localcord{k_{\bm{X}}}(\bm{X}_{1:N-1}, \bm{X}^*_{1:M-1}), \quad\text{and} \nonumber \\ 
& \quad \localcord{\bm{L}^*} = \localcord{k_{\bm{X}}}(\bm{X}^*_{1:M-1}, \bm{X}_{1:N-1})  (\localcord{\bm{K}_{\bm{X}}} + \localcord{\tilde{\bm{\Sigma}}_{\bm{X}}})^{-1}. \nonumber
\end{align}
We solve the optimization problem~\eqref{eq:hyperbolic-conditional-prediction} using Riemannian Adam~\cite{Becigneul19:RiemannianAdaptiveOpt}. To speed up convergence, we initialize the trajectory segments of $\bm{X}^*$ as hyperbolic geodesics.

\subsection{Pullback-metric geodesic motion generation}
\label{subsec:pullback}
From a Riemannian perspective, motion generation can be cast as geodesics computation~\cite{Jaquier2024:GPHLVM}. However, geodesics derived from the intrinsic metric of the hyperbolic manifold often risk traversing regions of low data density, where the model's uncertainty is high~\cite{Augenstein25:Hyperbolic}. To overcome this, we endow the latent space with a pullback metric induced by the stochastic mapping of the \ac{gp} (see Sec.\ref{sec:background}).
By computing pullback-metric geodesics, we can generate paths confined to the learned data manifold, ensuring that the resulting trajectories comply with the data distribution. Formally, the immersion $f$ represented by the \ac{gp} is stochastic, i.e., $f$ induces a distribution over its Jacobian $\bm{J}$, which itself induces a distribution over the pullback metric~\cite{Tosi14:RiemannianGPLVM}. 

In the Euclidean case, Tosi \etal~\cite{Tosi14:RiemannianGPLVM} assumed that the probability over $\bm{J}$ follows a Gaussian distribution, and has independent rows $\bm{J}_d \in \euclideanspace^{D_x}$, each with its own mean but shared covariance matrix. Thus, the Jacobian distribution is,
\begin{equation}
    p(\bm{J}) = \prod_{d=1}^{D_y} \mathcal{N}(\bm{J}_d \mid \bm{\mu}_{\bm{J}_d}, \bm{\Sigma}_{\bm{J}}) \, .
    \label{eq:EuclideanJacobianDistribution}
\end{equation}
The corresponding metric tensor follows a non-central Wishart distribution~\cite{Anderson46:Wishart},
${p(\euclpullbackmetric_{\bm{x}}) = \mathcal{\bm{W}}_{D_x}(D_y, \bm{\Sigma}_{\bm{J}}, \mathbb{E}[\bm{J}]^\trsp \mathbb{E}[\bm{J}])}$,
from which the expected metric tensor is computed as
$\mathbb{E}[\euclpullbackmetric_{\bm{x}}] = \mathbb{E}[\bm{J}]^\trsp \mathbb{E}[\bm{J}] + D_y \bm{\Sigma}_{\bm{J}}$.
In the hyperbolic case, Augenstein \etal~\cite{Augenstein25:Hyperbolic} follow a similar strategy. In this context, our GPHDM is a stochastic mapping of the form $f: \lorentz{D_x} \rightarrow \euclideanspace^{D_y}$ (as the GPLHVM in~\cite{Jaquier2024:GPHLVM}), whose Jacobian domain is restricted to the hyperbolic tangent space. In this case, the Jacobian distribution is, with restricted domain,
\begin{equation}
    p(\bm{J}^{\mathcal{L}}) = \prod_{d=1}^{D_y} \gaussiandist{\bm{J}_d^{\mathcal{L}}}{ \bm{\mu}_{\bm{J}_d}}{ \bm{\Sigma}_{\bm{J}}} \, .
    \label{eq:RiemannianJacobianDistribution}
\end{equation}
As $\euclpullbackmetric_{\bm{x}}$, the metric tensor $\lorentzpullbackmetric_{\bm{x}}$ follows a non-central Wishart distribution ${p(\lorentzpullbackmetric_{\bm{x}}) = \mathcal{\bm{W}}_{D_x}(D_y, \bm{\Sigma}_J, \mathbb{E}[\bm{J}^{\mathcal{L}}]^\trsp \mathbb{E}[\bm{J}^{\mathcal{L}}])}$, leading to the expected metric~\cite{Augenstein25:Hyperbolic}, 
\begin{equation}
    \mathbb{E}[\lorentzpullbackmetric_{\bm{x}^*}] = \bm{\mu}_{\bm{J}}^\trsp \bm{\mu}_{\bm{J}} + D_y \bm{\Sigma}_{\bm{J}}\, .
    \label{eq:hyperbolic-pullback-metric-tensor}
\end{equation} 

\begin{figure*}[tbp]
    \centering
    \begin{subfigure}[b]{0.235\linewidth}
        \centering
        \includegraphics[trim={2.5cm 1.8cm 0.5cm 0.2cm},clip,width=0.85\linewidth]{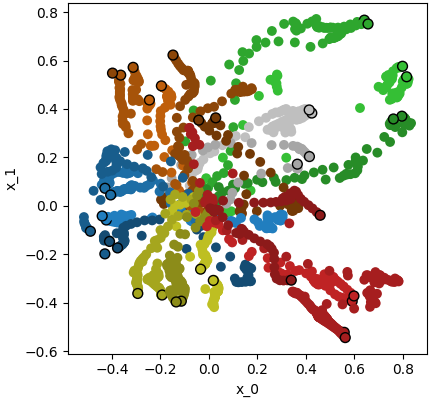}
        \includegraphics[width=0.95\linewidth]{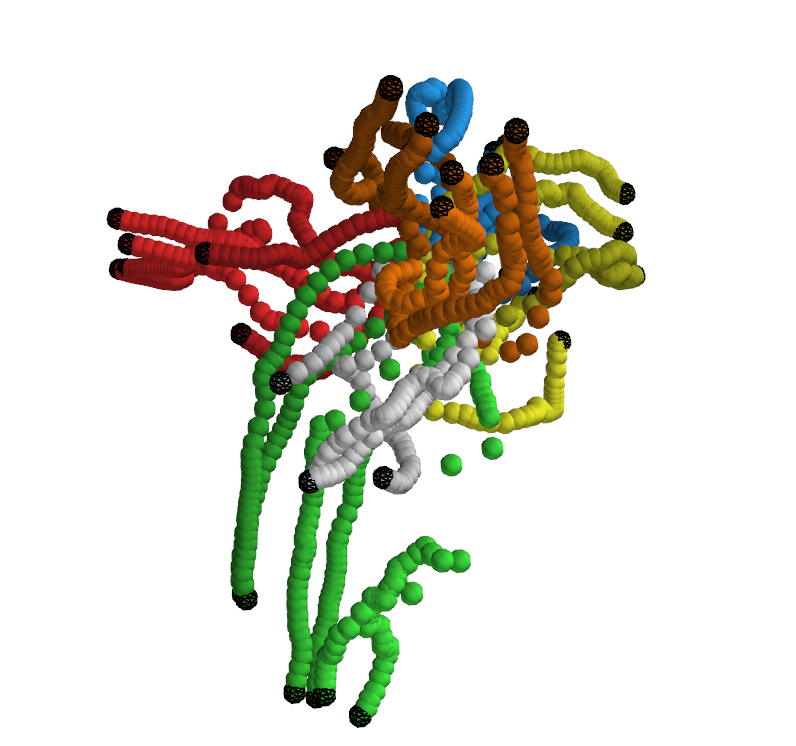}
        \caption{\ac{gplvm}}
        \label{subfig:gplvm}
    \end{subfigure}
    \hspace{-15pt}
    \begin{subfigure}[b]{0.235\linewidth}
        \centering
        \includegraphics[trim={2.5cm 1.8cm 0.5cm 0.2cm},clip,width=0.9\linewidth]{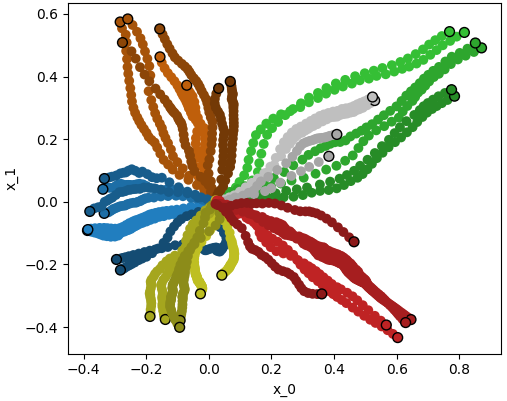}
        \includegraphics[width=0.95\linewidth]{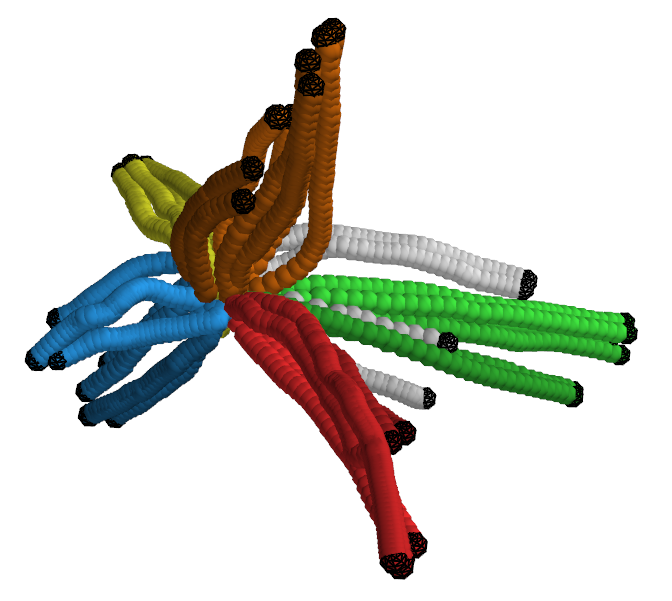}
        \caption{\ac{gpdm}}
        \label{subfig:gpdm}
    \end{subfigure}    
    \begin{subfigure}[b]{0.235\linewidth}
        \centering
        \includegraphics[width=0.85\linewidth]{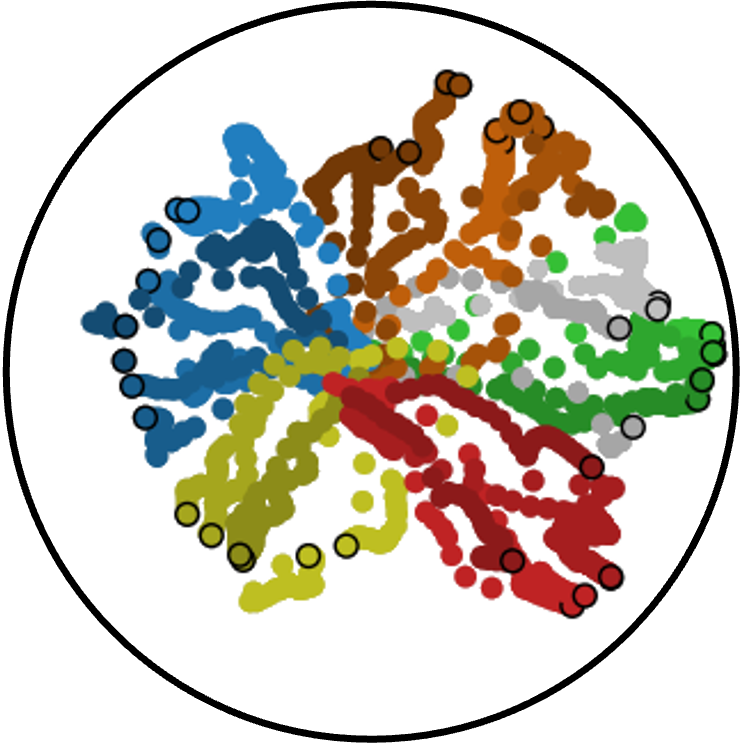}
        \vspace{0.1cm}
        
        \includegraphics[width=0.85\linewidth]{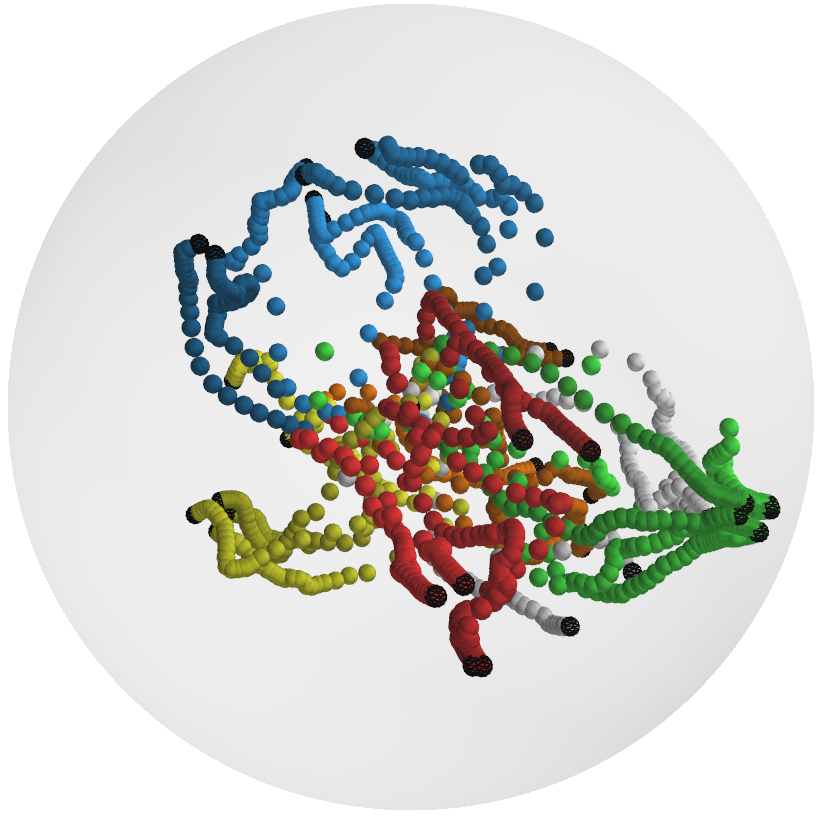}
        \caption{\ac{gphlvm}}
        \label{subfig:gphlvm}
    \end{subfigure}
    \begin{subfigure}[b]{0.235\linewidth}
        \centering
        \includegraphics[width=0.85\linewidth]{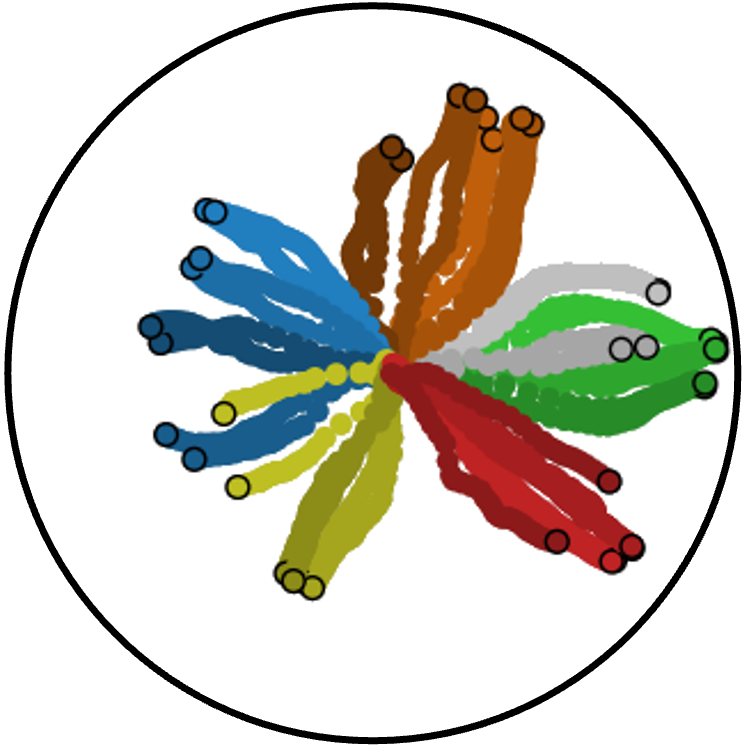}
        \vspace{0.1cm}
        
        \includegraphics[width=0.85\linewidth]{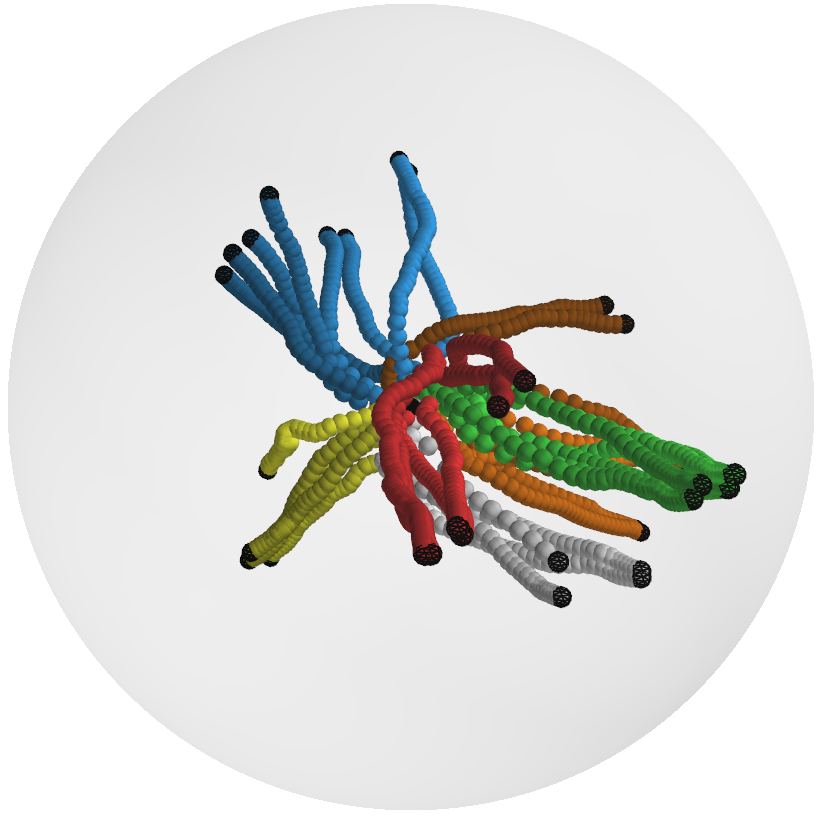}
        \caption{\ac{gphdm}}
        \label{subfig:gphdm}
    \end{subfigure}      
    \caption{Embeddings of hand grasps colored according to the grasp class of the last trajectory point with colors matching those of Fig.~\ref{fig:teaser}. The \textbf{top} and \textbf{bottom} rows show $2$- and $3$-dimensional latent spaces, respectively.}
    \label{fig:models-comparison}
    \vspace{-0.5cm}
\end{figure*} 

The pullback metric~\eqref{eq:hyperbolic-pullback-metric-tensor} allows us to compute geodesics that reflect the geometry of both the intrinsic hyperbolic space and the learned data distribution. These geodesics are computed by minimizing the curve length or, equivalently, the curve energy $E$ with respect to the pullback metric. For a geodesic discretized into a sequence of $M$ points  $\bm{x}_i\in \lorentz{D_x}$, this problem reduces to minimizing,
\begin{align}
    \label{eq:curve_energy}
    E = \sum_{i=0}^{M-2} \bm{v}_i^\trsp \lorentzpullbackmetric_{\bm{x}_i} \bm{v}_i \, , \quad \text{with} \quad \bm{v}_i = \text{Log}_{\bm{x}_i}({\bm{x}_{i+1}}).
\end{align}
Following~\cite{Augenstein25:Hyperbolic}, we regularize the objective~\eqref{eq:curve_energy} with the spline energy $E_{\text{spline}} \approx \sum_{i=1}^{M-2} \hypedist{\bm{x}_i}{\bar{\bm{x}}_i}^2$, 
where ${\bar{\bm{x}}_i = \expmap{\bm{x}_{i-1}}{\frac{1}{2} \, \logmap{\bm{x}_{i-1}}{\bm{x}_{i+1}} }}$ represents the geodesic midpoint between $\bm{x}_{i-1}$ and $\bm{x}_{i+1}$. This regularization term prevents uneven spacing of the discrete points along the geodesics. Finally, our geodesic optimization problem is, 
\begin{equation}
\label{eq:loss_energy_regularized}
\min_{\bm{x}_0, ..., \bm{x}_{M-1}} E + \lambda E_{\text{spline}}, 
\end{equation}
with $\lambda$ balancing the regularization. As for the recursive motion generation in Sec.\ref{subsec:recursive_motion}, we employ  Riemannian Adam~\cite{Becigneul19:RiemannianAdaptiveOpt} to solve our geodesic optimization problem.

\section{EXPERIMENTS}
\label{sec:evaluation}
We test the proposed \ac{gphdm} and dynamic motion generation strategies on data associated with a hand grasp taxonomy~\cite{Stival19:HumanGraspTaxonomy} that organizes common grasp types~\cite{Feix16:GRASPtaxonomy} into a tree structure based on their muscular and kinematic properties (see Fig.~\ref{fig:teaser}). 
We use a dataset from the KIT whole-body motion database~\cite{Mandery16:KITmotionDatabase} consisting of $38$ motions of $19$ common grasp types obtained from recordings of humans grasping different objects. For each motion, a subject (ID $2122$ or $2123$) reaches out from an initial resting pose to grasp an object placed on a table. The dataset consists of trajectories $\bm{Y} \in \euclideanspace^{N \times 24}$ representing the temporal evolution of the $24$ degrees of freedom of the wrist and fingers.
We preprocessed the recorded data by: (1) Applying a low-pass filter to remove high-frequency noise; (2) Trimming the start and end of each trajectory to keep only the motion from the initial resting pose to the grasp completion, that we define as the time when the grasped object is first moved by the human; (3) Subsampling the trajectories; and (4) Centering the data to allow for the use of a zero mean function in the \acp{gplvm}. After preprocessing, each trajectory is composed of 30 to 40 data points for a total of $N = 1321$ data points.

\subsection{Hyperbolic Embeddings of Taxonomy-structured Motions}
\label{subsec:experiment-gphdm}
We embed the trajectories associated to the aforementioned grasp taxonomy into $2$- and $3$-dimensional hyperbolic and Euclidean spaces using back-constrained \ac{gphlvm}, \ac{gphdm}, \ac{gplvm}, and \ac{gpdm}. We initialize the latent variables of all models by minimizing the stress~\eqref{eq:stressLoss} of the start and end points of each trajectory with respect to their associated taxonomy nodes, using the hyperbolic and Euclidean distance for the respective models. Then, we initialize the intermediate points of each trajectory as equally spaced along the hyperbolic geodesic connecting the obtained start and end points. \looseness-1

Fig.~\ref{fig:models-comparison} shows the obtained latent spaces. We observe that, for all models, the hand's initial resting pose is located near the latent space origin, from where the embedded trajectories progress outwards until the final grasp. 
Due to the stress prior, final grasps are organized in the latent space according to the taxonomy with grasps of each node grouped together in the latent space. 
Importantly, the trajectories appear scattered in the latent spaces of both \ac{gplvm} (Fig.~\ref{subfig:gplvm}) and \ac{gphlvm} (Fig.~\ref{subfig:gphlvm}). In contrast, the dynamics priors of the \ac{gpdm} (Fig.~\ref{subfig:gpdm}) and \ac{gphdm} (Fig.~\ref{subfig:gphdm}) effectively lead to smooth trajectories between the resting pose and the final grasps.

A quantitative evaluation of the models is presented in Table~\ref{tab:models-comparisons}. The stress reported in the first columns shows that all hyperbolic models better capture the taxonomy structure that their Euclidean counterparts. The difference is more prominent in the $2$-dimensional case, consistent with the ability of hyperbolic geometry to embed tree-like structures in low-dimensional spaces. In general, we observed similar stress values between the \ac{gphlvm} and \ac{gphdm}.
The second column of Table~\ref{tab:models-comparisons} reports the mean squared jerk (MSJ), which quantifies the smoothness of the latent trajectories. The MSJ of a single trajectory is given by, 
\begin{align}
    &\, \text{MSJ}(\bm{x}_1, ..., \bm{x}_{N}) = \dfrac{1}{N-4} \sum_{t=4}^N \vert \Delta^2 v_t \vert^2, \quad \text{with} \\
    &\, \Delta^2 v_t = v_t - 2 v_{t-1} + v_{t-2} \quad \text{and} \quad v_t = \manifolddist{\bm{x}_t}{ \bm{x}_{t-1}} , \nonumber
\end{align}
where $d_\manifold$ is the hyperbolic or Euclidean distance depending on the latent space geometry, $v_t$ is a discrete approximation of the velocity, and $\Delta^2$ is the finite difference operator.
We observe a prominent MSJ reduction for the \ac{gphdm} and \ac{gpdm} compared to the \ac{gphlvm} and \ac{gplvm}, indicating that both hyperbolic and Euclidean dynamics priors effectively promote smooth trajectories. Moreover, the \ac{gphdm} achieves the lowest MSJ among all models for both latent space dimensionality. Finally, the last column of Table~\ref{tab:models-comparisons} reports the mean squared error (MSE) of the decoded trajectories with respect to the training data. All models achieve similar MSE with $3$-dimensional latent spaces leading to lower errors.
In summary, the \ac{gphdm} inherits the superior taxonomy preservation of the \ac{gphlvm}, while simultaneously preserving the trajectory dynamical structure akin to the \ac{gpdm}.

\begin{table}[t]
    \caption{Average stress, mean squared jerk (MSJ), and reconstruction mean squared error (MSE) per model and geometry.}
    \vspace{-0.2cm}
    \label{tab:models-comparisons}
    \begin{center}
    \begin{tabular}{cccc}
        \toprule
        \textbf{Model} & \textbf{Stress} $\downarrow$& \textbf{MSJ} ($\times 100$) $\downarrow$ & \textbf{MSE} ($\times 100$) $\downarrow$ \\
        \midrule
        \ac{gplvm} $\euclideanspace^2$ & $0.21\pm 0.43$ &	$1.87\pm 6.15$ & $\bm{0.50\pm 1.28}$ \\
        \ac{gpdm} $\euclideanspace^2$ & $0.18\pm 0.47$ &	$\underline{0.04\pm 0.04}$ & $2.11\pm 5.13$ \\
        \ac{gphlvm} $\lorentz{2}$ & $\bm{0.09\pm 0.19}$ & $2.53 \pm 10.30$	& $\underline{1.26 \pm 2.70}$\\
        \ac{gphdm} $\lorentz{2}$ & $\underline{0.10\pm0.26}$ & $\bm{0.03\pm 0.02}$ & $2.03\pm4.40$\\
        \midrule
        \ac{gplvm} $\euclideanspace^3$ & $0.14\pm 0.21$ & $0.58 \pm 2.06$ & $\bm{0.29 \pm 0.76}$ \\
        \ac{gpdm} $\euclideanspace^3$ & $0.11\pm 0.21$ & $\underline{0.02\pm 0.04}$ & $\underline{0.31\pm 0.99}$ \\
        \ac{gphlvm} $\lorentz{3}$ & $\underline{0.09 \pm 0.13}$ & $1.68 \pm 7.47$ & $0.32\pm 0.95$\\
        \ac{gphdm} $\lorentz{3}$ & $\bm{0.07\pm 0.10}$ & $\bm{0.02\pm 0.02}$ & $0.32\pm 1.04$\\
        \bottomrule
    \end{tabular}
    \end{center}
    \vspace{-0.6cm}
\end{table}

\subsection{Dynamic Motion Generation}
\begin{figure}[tbp]
    \centering
    \includegraphics[width=0.39\linewidth]{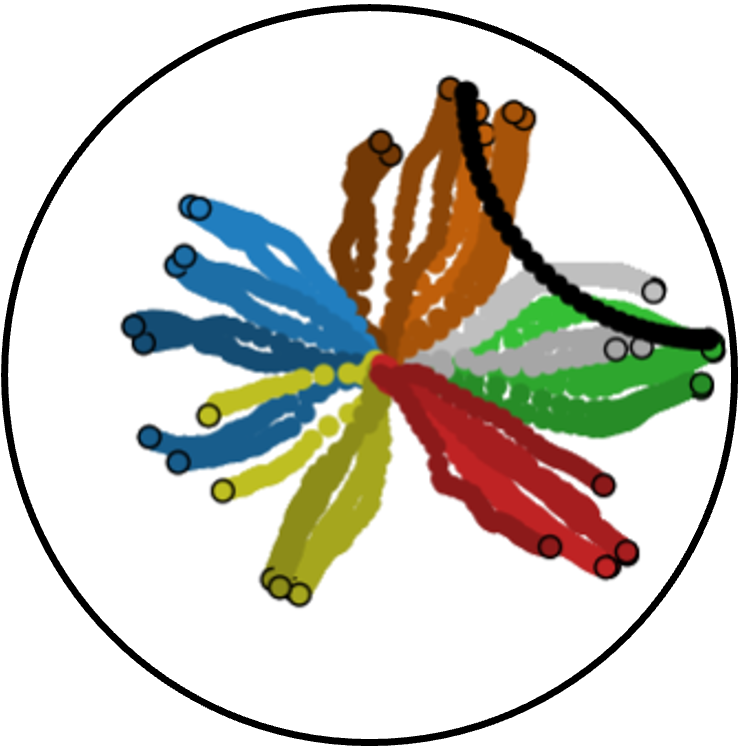}
    \includegraphics[width=0.42\linewidth]{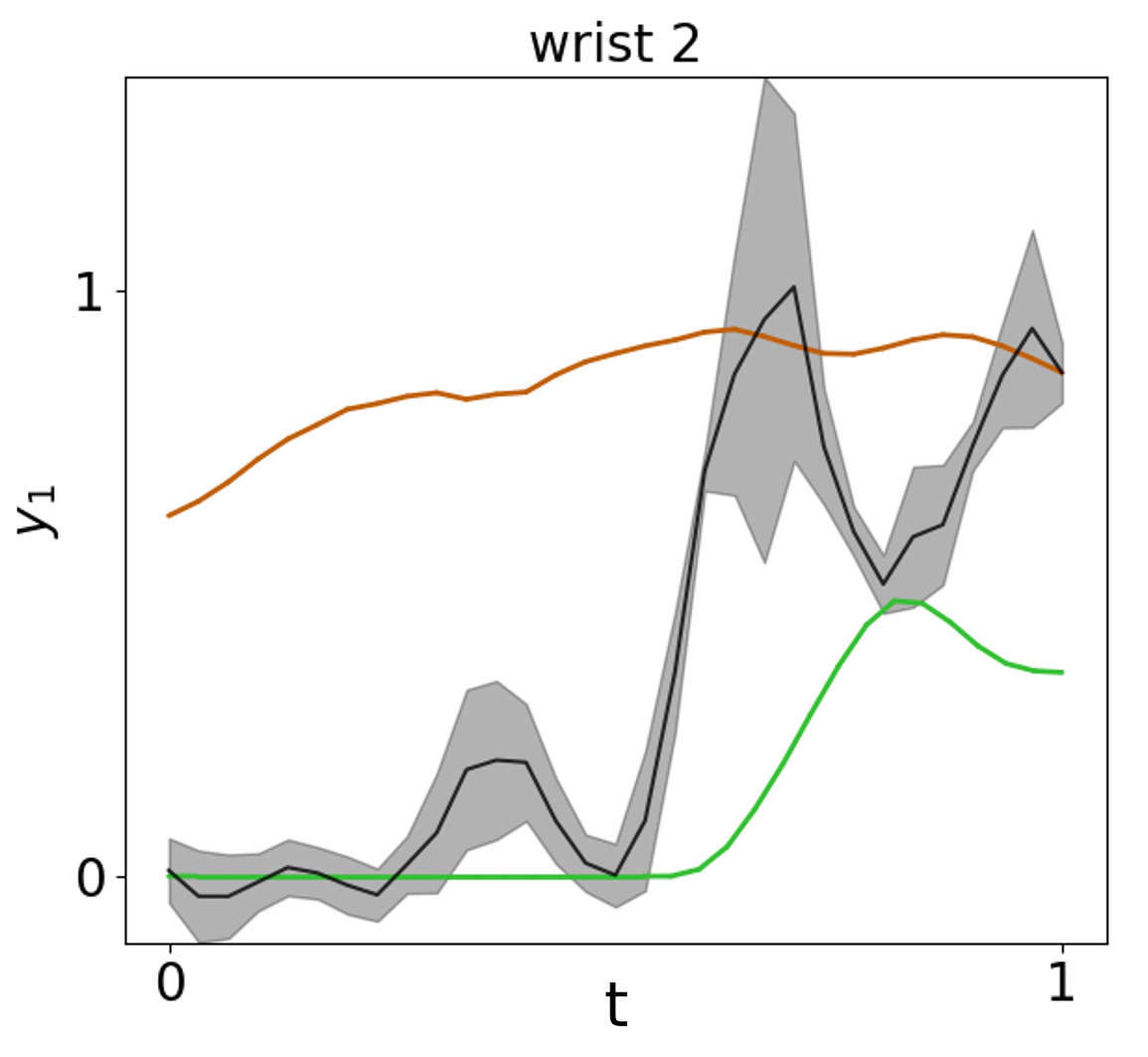}
    \vspace{0.1cm}

    \includegraphics[width=0.39\linewidth]{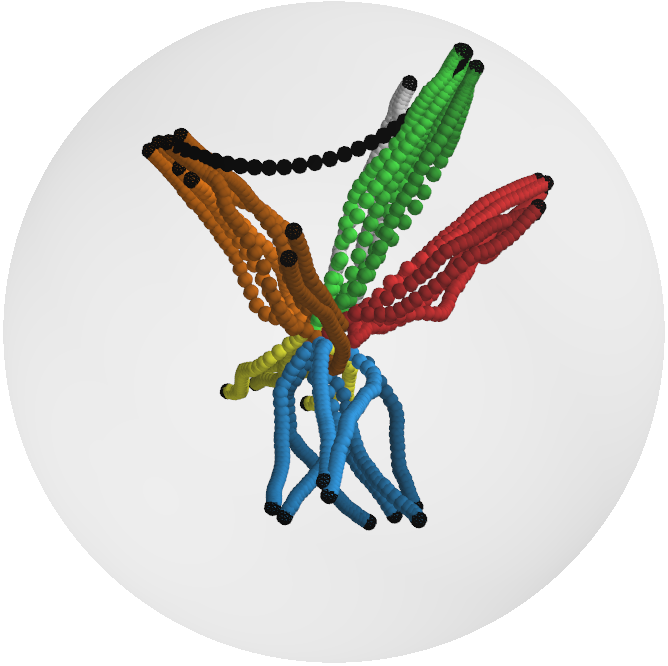}
    \includegraphics[width=0.42\linewidth]{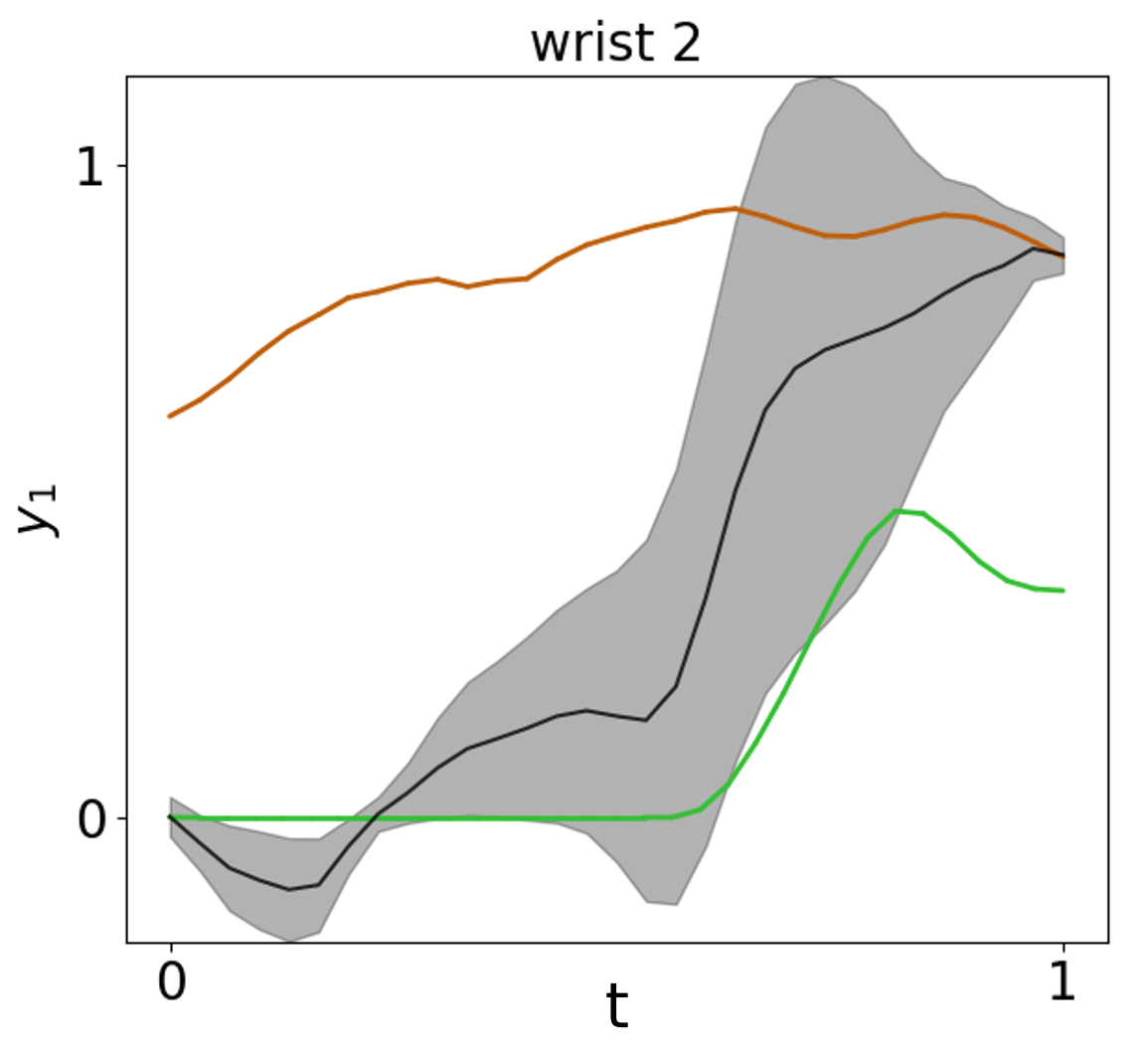}
    \caption{\textbf{Left}: Embeddings of hand grasps and hyperbolic geodesics (\blackline) from a lateral (\middlegreencircle) to a stick (\marrooncircle) grasp. \textbf{Right}: Representative dimension of the probabilistic hand motion prediction for the geodesics (\blackline) with mean and uncertainty, along with training trajectories for the lateral (reversed) (\middlegreenline) and stick (\marroonline) grasps. The \textbf{top} and \textbf{bottom} rows show $2$- and $3$-dimensional latent spaces.}
    \label{fig:2D-gphdm-geodesic}
    \vspace{-0.3cm}
\end{figure}

We evaluate the three different approaches to generate novel motions by decoding latent space trajectories. 
Fig.~\ref{fig:2D-gphdm-geodesic} shows the hyperbolic geodesics interpolating from a lateral to a stick grasp in the latent space of $2$- and $3$-dimensional \acp{gphdm}, akin to the approach proposed in~\cite{Jaquier2024:GPHLVM}, alongside one exemplary dimension of the decoded hand motions.
We observe that the $2$-dimensional geodesic does not follow the structure of the training data and crosses the training trajectories, thus resulting in jerky hand motions. Moreover, the $3$-dimensional geodesic predominantly passes through data-sparse regions, resulting in high-uncertainty motion predictions that revert to the non-informative mean. In both cases, the geodesics result in physically impractical motions that fail to capture the underlying dynamics of the training data.

\begin{figure}[tbp]
    \centering
    \includegraphics[width=0.44\linewidth]{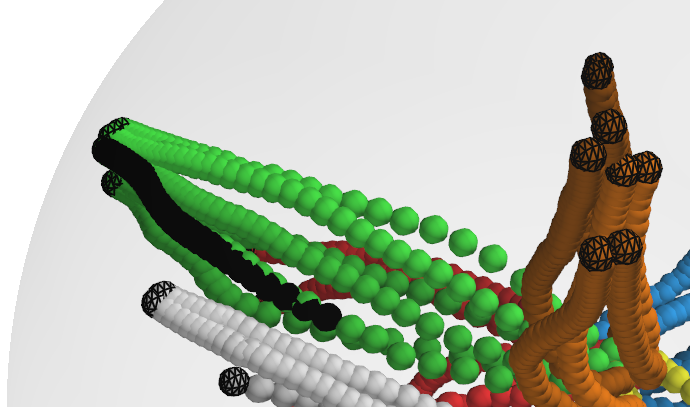}
    \hspace{0.05cm}
    \includegraphics[trim={0.2cm 11.8cm 5.5cm 3.2cm},clip,width=0.44\linewidth]{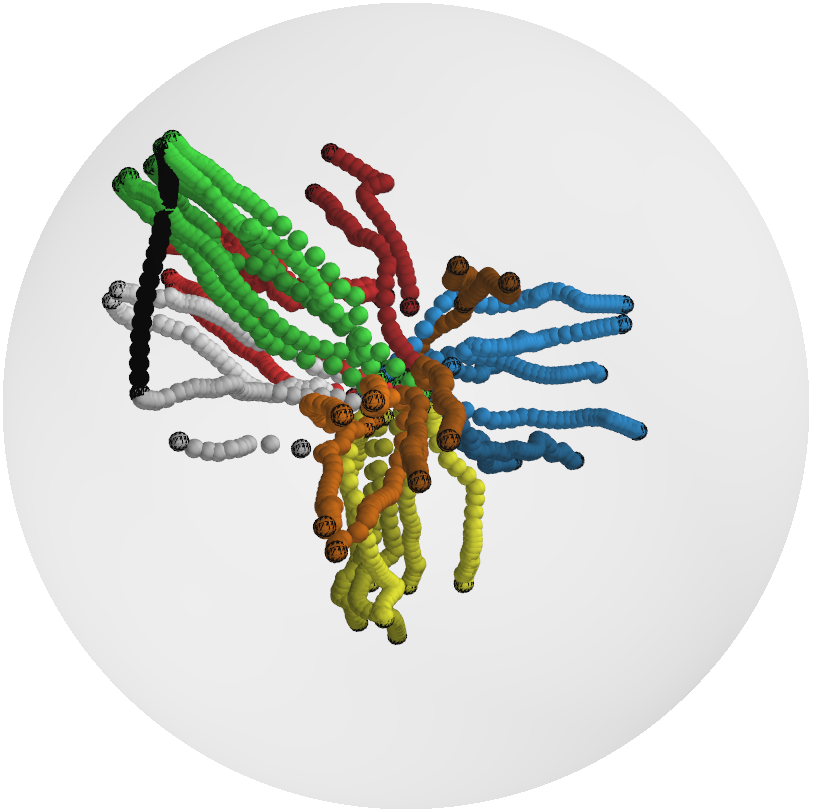}
    \caption{Latent trajectories (\blackline) obtained via recursive motion generation in a 3-dimensional \ac{gphdm} (zoomed-in). \textbf{Left}: Mean prediction towards a lateral grasp (\middlegreencircle). \textbf{Right}: Conditional prediction from an index finger extension (\lightgraycircle) to a lateral (\middlegreencircle) grasp.}
    \label{fig:mean-prediction}
    \vspace{-0.6cm}
\end{figure}

\begin{figure}[tbp]
    \centering
    \includegraphics[width=0.36\linewidth]{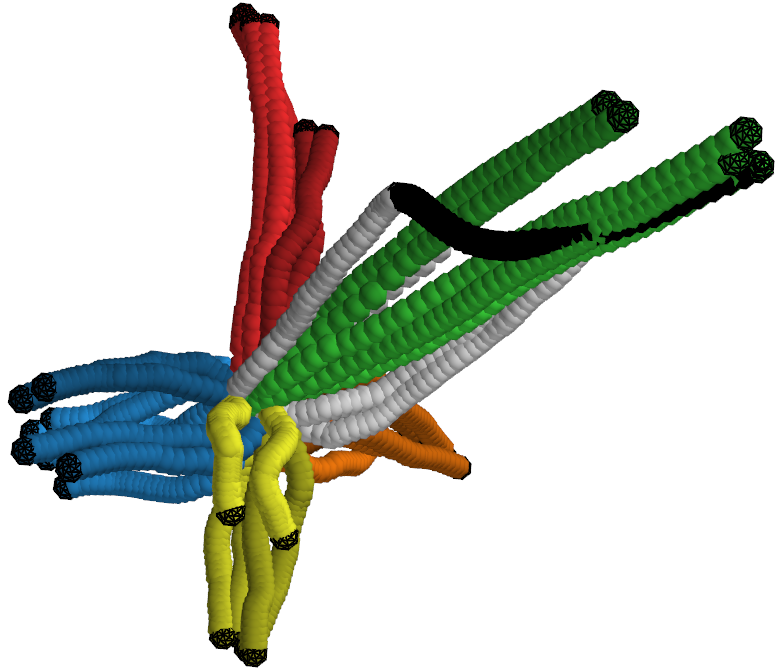}
    \includegraphics[width=0.36\linewidth]{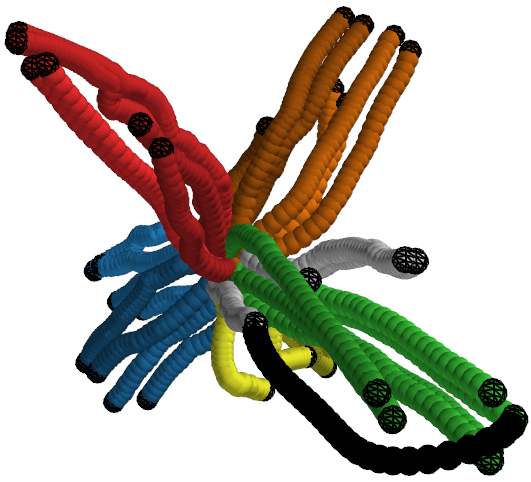}  
    \caption{Illustration of the directionality induced by the \ac{gpdm}'s Markov prior on latent trajectories obtained via conditional predictions. The outward transition (\textbf{left}) from an index finger extension (\lightgraycircle) to a lateral (\middlegreencircle) grasp follows a training trajectory, while the reverse transition (\textbf{right}) avoids the training data.}
    \label{fig:conditional-optimization}
    \vspace{-0.3cm}
\end{figure}

Fig.~\ref{fig:mean-prediction}-\emph{left} displays a latent trajectory obtained via the mean prediction approach introduced in Sec.~\ref{subsec:recursive_motion}. The latent trajectory is initialized in the middle of a training trajectory leading to a lateral grasp. We observe that the dynamics prior induces an outward direction, resulting in motion predictions that closely follow the training trajectory. However, as previously discussed, the mean prediction approach does not allow us to specify a desired goal point for the latent trajectory. 
Fig.~\ref{fig:mean-prediction}-\emph{right} shows a latent trajectory obtained via the conditional prediction approach of Sec.~\ref{subsec:recursive_motion} with start and goal points set as an index finger extension and a lateral grasp, respectively. In contrast to geodesics, the obtained trajectory incorporates the hyperbolic dynamics prior to transition between the two grasps. 
However, the conditional predictions follow the Markov assumption embedded in the hyperbolic dynamics prior, which induces a sense of directionality in the latent trajectories, as illustrated in Fig.~\ref{fig:conditional-optimization}. To avoid this issue, we trained the \ac{gphdm} of Fig.~\ref{fig:mean-prediction}-\emph{right} on an augmented dataset by including reverse training motions, i.e., motions from the final grasp onto the initial resting pose. Importantly, the recursive motion generation strategies do not prevent latent trajectories to traverse data-sparse regions, and thus often lead to motion predictions reverting to the mean and featuring high uncertainty.

\begin{figure}[tbp]
    \centering
    \includegraphics[width=0.43\linewidth]{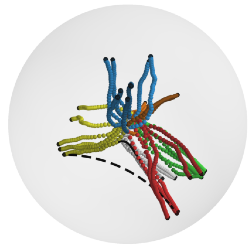}
    \hspace{0.2cm}
    \includegraphics[width=0.27\linewidth]{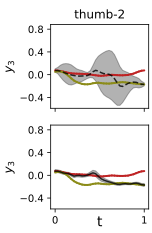}

    \includegraphics[width=0.95\linewidth]{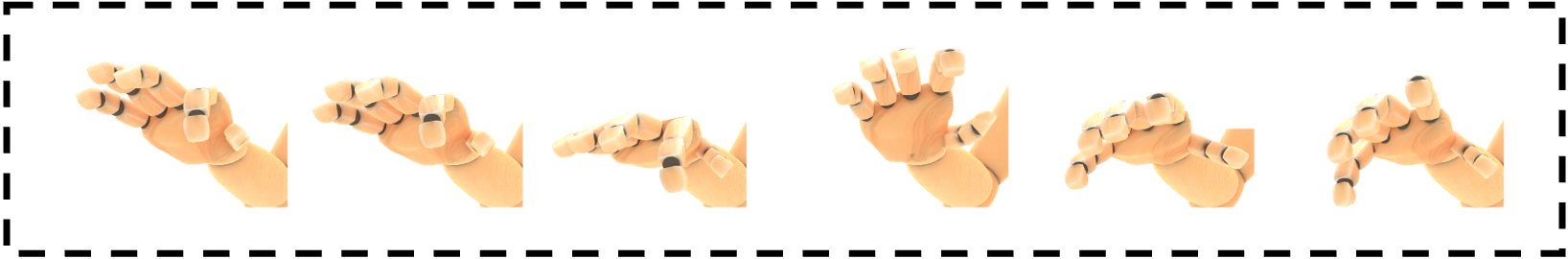}
    \includegraphics[width=0.95\linewidth]{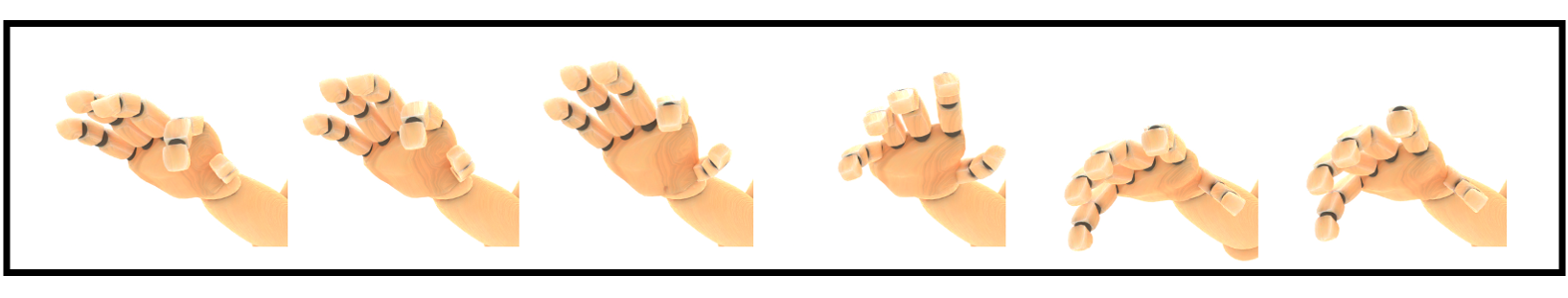}
    \caption{\textbf{Top left}: 3-dimensional embeddings of hand grasps via a \ac{gphdm} with a hyperbolic (\blackdashedline) and a pullback (\blackline) geodesic from a ring (\crimsoncircle) to a spherical (\olivecircle) grasp.
    \textbf{Top right}: Representative dimension of the probabilistic hand motion prediction with mean and uncertainty along with training trajectories for the spherical (\oliveline) and ring (reversed) (\crimsonline) grasps. \textbf{Bottom}: Generated hand motions from the decoded geodesics.}
    \vspace{-0.6cm}
    \label{fig:hand_grasps_3D_pullback}
\end{figure}

Fig.~\ref{fig:hand_grasps_3D_pullback} shows a latent trajectory obtained as a geodesic with respect to the \ac{gphdm}'s pullback metric introduced in Sec.~\ref{subsec:pullback}. The corresponding hyperbolic geodesic (akin to Fig.~\ref{fig:2D-gphdm-geodesic}) is depicted as a reference. We observe that, in contrast to the hyperbolic geodesic and conditionally-optimized latent trajectories, the pullback geodesic closely adheres to the data support. Thus, the pullback-metric geodesic results in low-uncertainty motion predictions. As shown at the bottom of Fig.~\ref{fig:hand_grasps_3D_pullback}, the decoded hyperbolic geodesic leads to hand motions displaying large deviations from the start ring and goal spherical grasps. In contrast, by capturing the underlying motion dynamics and adhering to the training data, the decoded pullback geodesic produces physically-plausible motions with little deviation from the start and goal grasps. 

\section{CONCLUSIONS}
This paper proposed the \ac{gphdm}, a model that leverages hyperbolic geometry, human-designed taxonomy structures, and dynamics priors as inductive biases to learn latent spaces that preserve the hierarchical structure and temporal dynamics of human motions. We showed that these three forms of inductive biases are essential to learn taxonomy-aware dynamically-consistent latent spaces. Moreover, we introduced three novel mechanisms for generating taxonomy-aware and physically-consistent motions. Our results showed that trajectories obtained as geodesics on the pullback metric of the learned model produced low-uncertainty, physically-consistent motions that capture hierarchical structure and temporal dynamics of the motion data. 

\bibliographystyle{IEEEtran}

\bibliography{References}

@string{T-RO			= "{IEEE} T-RO"}

@string{PAMI 		= "{IEEE} PAMI"}

@string{IROS    		= "{IEEE/RSJ} {IROS}"}

@string{R:SS    		= "{R:SS}"}

@string{ICML   		= "{ICML}"}

@string{NeurIPS 		= "{NeurIPS}"}

@string{CVPR 		= "{CVPR}"}

@string{CoRL 		= "{CoRL}"}

@string{AISTATS 	= "{AISTATS}"}

@string{ICLR 	    = "{ICLR}"}

@string{UAI 	    = "{UAI}"}

@InProceedings{Jaquier2024:GPHLVM,
	Title       = {Bringing motion taxonomies to continuous domains via {GPLVM} on hyperbolic manifolds},
	Author      = {Jaquier, No\'emie and Rozo, Leonel and Gonz\'alez-Duque, Miguel and Borovitskiy, Viacheslav and Asfour, Tamim},
	Booktitle   = ICML,
	Year        = {2024},
	Pages      = {},
}

@inproceedings{Jaquier21:GaBOMatern,
	author = {Jaquier, Noémie and Borovitskiy, Viacheslav and Smolensky, Andrei and Terenin, Alexander and Asfour, Tamim and Rozo, Leonel}, 
	title = {Geometry-aware {B}ayesian Optimization in Robotics using {R}iemannian {M}at\'ern Kernels},
	booktitle = CoRL,
	year = {2021},
    nourl = {https://openreview.net/forum?id=ovRdr3FOIIm},
	pages = {}
}

@ARTICLE{Wang08:GPDM,  
    author = {Wang, Jack M. and Fleet, David J. and Hertzmann, Aaron},  
    journal = PAMI,   
    title={Gaussian Process Dynamical Models for Human Motion},   
    year={2008},  
    volume={30},  
    number={2},  
    pages={283-298},  
    doi={10.1109/TPAMI.2007.1167}
}

@inproceedings{Lawrence03:GPLVM,
 author = {Lawrence, Neil D.},
 booktitle = NeurIPS,
 title = {Gaussian Process Latent Variable Models for Visualisation of High Dimensional Data},
 nourl = {https://proceedings.neurips.cc/paper/2003/file/9657c1fffd38824e5ab0472e022e577e-Paper.pdf},
 year = {2003}
}

@inproceedings{Urtasun08:TopologicalGPLVM,
    author = {Urtasun, Raquel and Fleet, David J. and Geiger, Andreas and Popovi\'{c}, Jovan and Darrell, Trevor J. and Lawrence, Neil D.},
    title = {Topologically-Constrained Latent Variable Models},
    year = {2008},
    doi = {10.1145/1390156.1390292},
    booktitle = ICML,
    pages = {1080–1087}
}

@inproceedings{Lawrence06:BackConstrGPLVM,
    author = {Lawrence, Neil D. and Qui\~{n}onero-Candela, Joaquin},
    title = {Local Distance Preservation in the {GP-LVM} through Back Constraints},
    year = {2006},
    doi = {10.1145/1143844.1143909},
    booktitle = ICML,
    pages = {513–520}
}

@article{Borras17:WholeBodyTaxonomy,
    author = {Júlia Borr\`as  and Christian Mandery  and Tamim Asfour },
    title = {A whole-body support pose taxonomy for multi-contact humanoid robot motions},
    journal = {Sci. Robot.},
    volume = {2},
    number = {13},
    year = {2017},
    doi = {10.1126/scirobotics.aaq0560}
}

@inproceedings{Nickel2017:Poincare,
	author = {Nickel, Maximillian and Kiela, Douwe},
	booktitle = NeurIPS,
	title = {Poincaré Embeddings for Learning Hierarchical Representations},
	nourl = {https://arxiv.org/abs/1705.08039},
	year = {2017}
}

@article{GrigoryanNoguchi98:HyperbolicHeatKernel,
    title = {The heat kernel on hyperbolic space},
	author = {Grigoryan, Alexander and Noguchi, Masakazu},
    journal = {Bull. Lond. Math. Soc.},
	fulljournal = {Bulletin of the London Mathematical Society},
	number = {6},
	pages = {643--650},
	volume = {30},
	year = {1998},
	doi = {10.1112/S0024609398004780}
}

@book{Lee18:RiemannManifold,
	author = {John Lee},
	title = {Introduction to {R}iemannian Manifolds},
	year = {2018},
	publisher = {Springer},
	doi = {10.1007/978-3-319-91755-9},
	noedition = {2nd}
}

@book{Ratcliffe19:HyperbolicManifold,
	author = {John G. Ratcliffe},
	title = {Foundations of Hyperbolic Manifolds},
	year = {2019},
	publisher = {Springer},
	doi = {10.1007/978-3-030-31597-9},
	noedition = {3rd}
}

@Book{Boumal22:RiemannOpt,
  title        = {An introduction to optimization on smooth manifolds},
  author       = {Boumal, Nicolas},
  publisher = {Cambridge University Press},
  year         = {2023},
  nourl          = {http://www.nicolasboumal.net/book},
}

@inproceedings{Paulius20:ManipulationTaxonomy,
  title = {A Motion Taxonomy for Manipulation Embedding},
  author = {Paulius, David and Eales, Nicholas and Sun, Yu},
  booktitle = R:SS,
  pages = {},
  year = {2020},
  nourl = {http://www.roboticsproceedings.org/rss16/p045.pdf}
}

@INPROCEEDINGS{Mandery16:LanguageWholeBody,  
    author = {Mandery, Christian and Borràs, Júlia and Jöchner, Mirjam and Asfour, Tamim},
    booktitle = IROS,   
    title={Using language models to generate whole-body multi-contact motions},   year={2016},  
    pages={5411-5418},  
    doi={10.1109/IROS.2016.7759796}
}

@INPROCEEDINGS{Romero10:SpatioTempGraspsGPLVM,  
    author = {Romero, Javier and Feix, Thomas and Kjellström, Hedvig and Kragic, Danica},  
    booktitle = IROS,   
    title = {Spatio-temporal modeling of grasping actions},   
    year = {2010},  
    pages = {2103-2108},  
    doi = {10.1109/IROS.2010.5650701}
}

@article{Stival19:HumanGraspTaxonomy,
    author = {Francesca Stival and Stefano Michieletto and Matteo Cognolato and Enrico Pagello and Henning Müller and Manfredo Atzori},
    title = {A quantitative taxonomy of human hand grasps},
    journal = {J. Neuroeng Rehabil.},
    fulljournal = {Journal of NeuroEngineering and Rehabilitation},
    volume = {16},
    number = {28},
    year = {2019},
    doi = {10.1186/s12984-019-0488-x}
}

@ARTICLE{Feix16:GRASPtaxonomy,  
    author = {Feix, Thomas and Romero, Javier and Schmiedmayer, Heinz-Bodo and Dollar, Aaron M. and Kragic, Danica},  
    journal = {IEEE Trans. on Human-Machine Systems},   
    title = {The {GRASP} Taxonomy of Human Grasp Types},   
    year = {2016},  
    volume = {46},  
    number = {1},  
    pages={66-77},  
    doi = {10.1109/THMS.2015.2470657}
}

@article{Kochurov20:geoopt,
    title={Geoopt: {R}iemannian Optimization in {PyTorch}},
    author={Max Kochurov and Rasul Karimov and Serge Kozlukov},
    year={2020},
    journal={arXiv preprint arXiv:2005.02819},
    nourl = {https://github.com/geoopt/geoopt}
}

@inproceedings{Becigneul19:RiemannianAdaptiveOpt, 
    author = {Bécigneul, Gary and Ganea, Octavian-Eugen}, 
    title = {{R}iemannian Adaptive Optimization Methods}, 
    booktitle = ICLR, 
    year = {2019},
    nourl = {https://openreview.net/pdf?id=r1eiqi09K7}
}

@inproceedings{Chami2020:TreesHyperbolic,
	author = {Chami, Ines and Gu, Albert and Chatziafratis, Vaggos and Ré, Christopher},
	booktitle = NeurIPS,
	title = {From Trees to Continuous Embeddings and Back: Hyperbolic Hierarchical Clustering},
	nourl = {https://proceedings.neurips.cc/paper/2020/file/ac10ec1ace51b2d973cd87973a98d3ab-Paper.pdf},
	nopages = {15065--15076},
	year = {2020}
}

@inproceedings{Tosi14:RiemannianGPLVM, 
year = {2014}, 
author = {Tosi, Alessandra and Hauberg, Søren and Vellido, Alfredo and Lawrence, Neil D.}, 
title = {Metrics for Probabilistic Geometries}, 
booktitle = UAI, 
pages = {}
}

@ARTICLE {Mandery16:KITmotionDatabase,
author = {Christian Mandery and \"Omer Terlemez and Martin Do and Nikolaus Vahrenkamp and Tamim Asfour},
title = {Unifying Representations and Large-Scale Whole-Body Motion Databases for Studying Human Motion},
pages = {796--809},
volume ={32},
number ={4},
journal =T-RO,
year = {2016},
doi={10.1109/TRO.2016.2572685},
}

@article{Poincare00:PoincareModel,
    author = {Henri Poincar{\'e}},
    title = {{Théorie des groupes fuchsiens}},
    volume = {1},
    journal = {Acta Mathematica},
    publisher = {Institut Mittag-Leffler},
    pages = {1 -- 62},
    year = {1900},
    nourl = {https://doi.org/10.1007/BF02592124}
}

@article{Reynolds93:Hyperboloid,
    author = {William F. Reynolds},
    title = {Hyperbolic Geometry on a Hyperboloid},
    fulljournal = {The American Mathematical Monthly},
    journal = {Am. Math. Mon.},
    volume = {100},
    number = {5},
    pages = {442--455},
    year = {1993},
    publisher = {Taylor \& Francis},
    nourl = {https://doi.org/10.1080/00029890.1993.11990430}
}

@article{Jansen09:Hyperboloid,
  title = {Abbildung der hyperbolischen Geometrie auf ein zweischaliges Hyperboloid},
  author = {Jansen, H.},
  journal = {Mitt. Math. Ges. Hamburg},
  fulljournal = {Mitteilungen der Mathematischen Gesellschaft in Hamburg},
  volume = {4},
  pages = {409--440},
  year = {1909}
}

@ARTICLE{Anderson46:Wishart,
  author={Anderson, T.W.},
  journal = {Ann. Math. Stat.},
  fulljournal={The Annals of Mathematical Statistics}, 
  title={The non-central {W}ishart distribution and certain problems of multivariate statistics}, 
  year={1946},
  volume={17},
  number={4},
  pages={409-431},
  nourl={https://www.jstor.org/stable/pdf/2236082.pdf}
}

@article{Augenstein25:Hyperbolic,
	Title       = {On Probabilistic Pullback Metrics on Latent Hyperbolic Manifolds},
	Author      = {Augenstein, Luis and Jaquier, Noémie and Asfour, Tamim and Rozo, Leonel},
	journal   = {arXiv preprint arXiv:2410.20850},
	Year        = {2024},
    volume      = {},
    issue       = {},
    pages       = {},
}

@inproceedings{Mallasto18:wrapped, 
  year      = {2018}, 
  author    = {Mallasto, Anton and Feragen, Aasa}, 
  title     = {Wrapped {G}aussian Process Regression on {R}iemannian Manifolds}, 
  booktitle = CVPR, 
  pages     = {5580--5588}, 
}

@incollection{Iida2016:BioinspiredRobots,
    author="Iida, Fumiya and Ijspeert, Auke Jan",
    noeditor="Siciliano, Bruno and Khatib, Oussama",
    title="Biologically Inspired Robotics",
    booktitle="Springer Handbook of Robotics",
    year="2016",
    nopublisher="Springer",
    pages="2015--2034",
    doi="10.1007/978-3-319-32552-1_75",
    nourl="https://doi.org/10.1007/978-3-319-32552-1_75"
}

@inproceedings{Rozo2025:Riemann2,
    title={Riemann{$^2$}: Learning {R}iemannian Submanifolds from {R}iemannian Data}, 
    author={Leonel Rozo and Miguel Gonz{\'a}lez-Duque and No{\'e}mie Jaquier and S{\o}ren Hauberg},
    booktitle=AISTATS,
    year={2025},
}

%%%%%%%%%%%%%%%%%%%%%%%%%%%%%%%%%%%%%%%%%%%%%%%%%%%%%%%%%%%%%%%%%%%%%%%%%%%%%%%%

\end{document}